\documentclass[10pt,twocolumn,letterpaper]{article}
 \usepackage{cvpr}              

\definecolor{cvprblue}{rgb}{0.21,0.49,0.74}
\usepackage[pagebackref,breaklinks,colorlinks,allcolors=cvprblue]{hyperref}
\usepackage[accsupp]{axessibility}  

\usepackage{multirow} 
\usepackage{booktabs} 
\usepackage{makecell} 
\def\paperID{11411} 
\def\confName{CVPR}
\def\confYear{2025}

\title{Detection-Friendly Nonuniformity Correction: A Union Framework for Infrared UAV Target Detection}

\author{
\fontsize{11.2}{13.5}\selectfont
Houzhang Fang$^1$$^\dagger$\thanks{Corresponding author.\;\;\; $^\dagger$Equal contribution.}, \;Xiaolin Wang$^1$$^\dagger$,\; Zengyang Li$^1$,\; Lu Wang$^1$,\; Qingshan Li$^1$,\; Yi Chang$^2$, \; Luxin Yan$^{2}$\smallskip\\
{\fontsize{11.15}{13.5}\selectfont $^{1}$Xidian\,University\quad $^{2}$Huazhong University of Science and Technology} \\
{\tt\footnotesize houzhangfang@xidian.edu.cn,\:wxl@stu.xidian.edu.cn,\:22009200694@stu.xidian.edu.cn,\:wanglu@xidian.edu.cn,}\vspace{-0.2em} \\ {\tt\footnotesize  qshli@mail.xidian.edu.cn,\:yichang@hust.edu.cn,\:yanluxin@hust.edu.cn}
}

\begin{document}
\maketitle
\begin{abstract}
Infrared unmanned aerial vehicle (UAV) images captured using thermal detectors are often affected by temperature-dependent low-frequency nonuniformity, which significantly reduces the contrast of the images. Detecting UAV targets under nonuniform conditions is crucial in UAV surveillance applications. Existing methods typically treat infrared nonuniformity correction (NUC) as a preprocessing step for detection, which leads to suboptimal performance. Balancing the two tasks while enhancing detection-beneficial information remains challenging. In this paper, we present a detection-friendly union framework, termed UniCD, that simultaneously addresses both infrared NUC and UAV target detection tasks in an end-to-end manner. We first model NUC as a small number of parameter estimation problem jointly driven by priors and data to generate detection-conducive images. Then, we incorporate a new auxiliary loss with target mask supervision into the backbone of the infrared UAV target detection network to strengthen target features while suppressing the background. To better balance correction and detection, we introduce a detection-guided self-supervised loss to reduce feature discrepancies between the two tasks, thereby enhancing detection robustness to varying nonuniformity levels. Additionally, we construct a new benchmark composed of 50,000 infrared images in various nonuniformity types, multi-scale UAV targets and rich backgrounds with target annotations, called IRBFD. Extensive experiments on IRBFD demonstrate that our UniCD is a robust union framework for NUC and UAV target detection while achieving real-time processing capabilities. Dataset can be available at \href{https://github.com/IVPLaboratory/UniCD}{https://github.com/IVPLaboratory/UniCD.} 
\end{abstract}

 
\vspace{-1em}
\section{Introduction} \label{sec:intro}
\begin{figure}[!t]
\centering 
\includegraphics[width=3.2in,keepaspectratio]{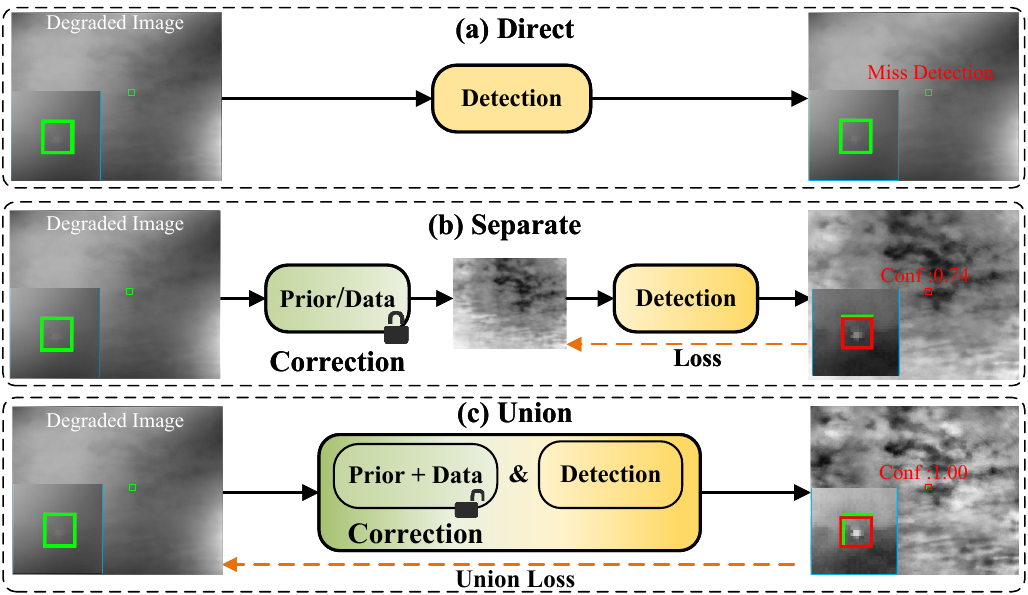}
\vspace{-0.6em} 
\caption{Three main categories of methods for UAV target detection in nonuniformity conditions. (a) Direct: detection models \cite{2023YOLOv8Jocher} are directly applied to nonuniformity degraded images. (b) Separate: correction model \cite{2016IPTLiu}  serves as a pre-processing step, correcting images before passing them to detectors \cite{2023YOLOv8Jocher}. (c) Union: correction and detection are processed simultaneously in a unified framework. Previous methods solely concentrates on optimizing one task. Our UniCD concurrently emphasizes the joint enhancement of correction quality and detection accuracy.} \vspace{-1.6em} 
\label{fig1} 
\end{figure}

Unmanned aerial vehicle (UAV) detection based on infrared imaging is an important perception technology for monitoring UAV in both day and night scenarios. However, the thermal radiation from the optical lens and the camera housing causes the acquired infrared UAV images to often suffer from the temperature-dependent low-frequency nonuniformity effects \cite{2010CVPRZheng,2016IPTLiu,2024TGRSXie,2024TGRSShi} (See the left column of \cref{fig1}). The optics-caused nonuniformity effect is also referred to as the bias field, which reduces the image contrast. Moreover, the infrared UAV targets typically have  weak features and complex backgrounds \cite{2023TIIFang,2023ACMMMFang}. 
Nonuniformity bias field further exacerbates the difficulty of UAV target detection. Infrared nonuniformity correction (NUC) and target detection have achieved significant advancements in recent years \cite{2019GRSLChang,2023TIIFang,2023ACMMMFang,2024CVPRLiu,2024TGRSShi,2023CVPRYing}. Previous methods focus on one aspect of the tasks and address the two tasks independently. As far as we know, no work considers the practical problem: infrared UAV target detection under the nonuniformity conditions.

To solve this problem, a simple strategy is to detect \cite{2023YOLOv8Jocher} UAV targets directly on the degraded bias field images (See \cref{fig1}(a)), which easily leads to miss detection due to the weakening of target information. Another typical approach is the correction-then-detection paradigm (See \cref{fig1}(b)). NUC methods \cite{2016IPTLiu,2022AOShi, 2024TGRSXie} are first adopted to remove the nonuniformity bias field, and the corrected images are then passed to the target detectors \cite{2023YOLOv8Jocher,2023TIIFang,2023ACMMMFang}. However, the existing NUC methods have limitations. Specifically, model-based NUC methods rely heavily on handcrafted features to model the images and bias fields, making them prone to overfitting the image content and thus struggle to handle complex or severely degraded bias fields \cite{2024TGRSXie}. Deep learning (DL)-based NUC methods depend on complex architectures and a large number of real input-output image pairs, which limits their practicality \cite{2019GRSLChang,2024TGRSShi}. Additionally, the NUC lacks supervision from the detection module to enhance detection-conducive information. Recently, joint methods for processing low-level images and high-level vision have already been proposed \cite{2022AAAILi,2022AAAILiu,2023TPAMILi}. However, they are primarily designed for object detection under adverse weather conditions.

To overcome the above limitations, in this paper, we propose a detection-friedly union framework, termed UniCD, that simultaneously tackles both infrared bias field correction and UAV target detection. On the one hand,  because of the spatially smooth nature of the bias field, we model it by the high-order 
bivariate polynomial \cite{2020TGRSLiu,2022AOShi,2024TGRSXie}, which can effectively fit the nonuniform bias field with different scales. 
As a result, accurately estimating the optimal polynomial coefficients is essential for ensuring the performance of bias field correction. 
In this work, we formulate bias field correction as a problem of predicting a small number of polynomial coefficients jointly driven by priors and data, which can be easily learned by a very lightweight network. The bias field has spatially continuous low-frequency characteristics, making transformer-based architecture well-suited for modeling this component. Additionally, we also integrate convolutional neural networks (CNNs) to capture the local details of the bias field.
By incorporating parametric prior modeling and low-dimensional data-driven prediction, our approach avoids dependence on handcrafted features and real input-output data pairs, significantly improving correction performance. 
On the other hand, 
existing DL-based detection methods mainly focus on designing complex model architectures for extracting features \cite{2022CVPRZhang,2023CVPRWLyu,2023TIIFang,2023ACMMMFang,2023CVPRYing,2024CVPRLiu,2024AAAIZhang}. We further introduce auxiliary loss with target mask supervision at different stages of the backbone in the infrared UAV target detection network without increasing computational complexity. Integrating this loss enhances the discriminative features of UAV targets while suppressing the background, thereby improving detection performance.

To balance correction and detection, we introduce a detection-guided self-supervised loss to reduce feature discrepancies between the correction and detection tasks. This loss enforces feature similarity between the corrected image and the reference image, both extracted by the detection backbone, thus ensuring high correction quality while enhancing features that are beneficial for detection. 
Our NUC model can also be flexibly integrated as a scalable module with existing detectors for infrared image bias field correction. Furthermore, we construct the bias field benchmark, IRBFD, consisting of 50,000 infrared images with varying nonuniformity types, multi-scale UAV targets, and rich backgrounds with target annotations, called IRBFD. Experimental results on the IRBFD demonstrate that UniCD outperforms state-of-the-art (SOTA) combined correction and detection methods in terms of precision and recall.

The contributions of this work can be summarized as:
\begin{itemize}
\item We propose a novel detection-friedly union framework, termed UniCD, that can simultaneously deal with NUC and infrared UAV target detection in an end-to-end manner. To the best of our knowledge, this is the first work to address both issues in a unified framework.
\item We for the first time model nonuniformity bias field correction as a problem of predicting a small number of hyperparameters jointly driven by priors and data, which can be easily performed with a very lightweight network.
\item We establish the first large benchmark called IRBFD to facilitate the research in the area of nonuniformity correction and infrared UAV target detection, which consists of 50,000 manually labeled infrared images with various nonuniformity levels, multi-scale UAV targets and rich backgrounds with target annotations.
\end{itemize}

\begin{figure*}[!t]
\centering 
\includegraphics[width=6.5in,keepaspectratio]{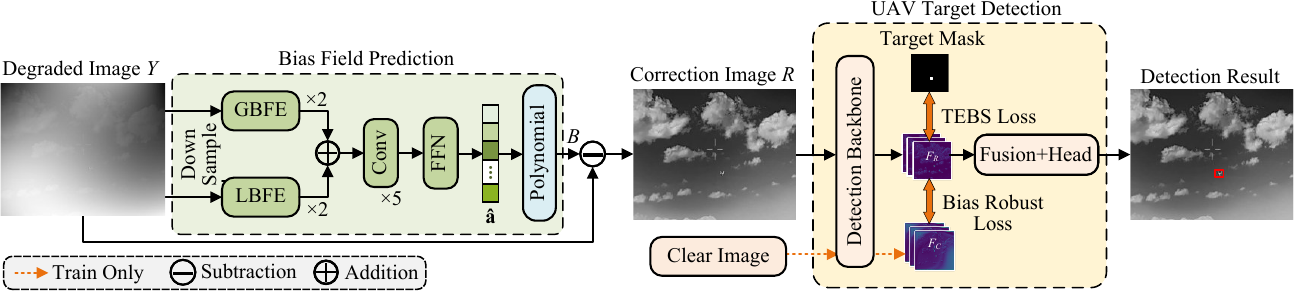}\vspace{-0.5em} 
\caption{Overview of the proposed UniCD. Our UniCD integrates a bias field prediction network with an infrared UAV target detection network. These two components are fused into a unified pipeline and trained end-to-end. The target enhancement and background suppression (TEBS) loss is introduced to enhance UAV target features while suppressing the background. The bias robust loss is employed to balance correction and detection. } 
\label{fig2} \vspace{-0.5em} 
\end{figure*}

\section{Related Work}  \label{sec:related}
\subsection{Nonuniformity Correction in Infrared Images}
NUC methods for removing the bias field are broadly divided into two main categories: model-driven methods and data-driven methods. Model-driven methods typically leverage the prior constraints of the bias field and the images to remove the nonuniformity effects \cite{2014OLCao,2016IPTLiu,2020TGRSLiu,2022AOShi,2024TGRSXie,2024OEWang}. However, these methods rely on handcrafted features with carefully tuned
hyper-parameters, limiting their practicality in real-world applications. Data-driven DL methods \cite{2019GRSLChang,2023IPTLiu,2024TGRSShi} have gained increasing attention. 
However, DL-based correction models are often complex and rely on large amounts of training data with real correction labels, which restricts their widespread application in practical scenarios \cite{2020TPAMIBridging,2020TIPYang,2023weatherproof}. In contrast, we formulate a novel lightweight correction model driven by both parametric priors and data, which converges more easily.

\subsection{Infrared UAV Target Detection}
In recent years, many methods have been developed to detect infrared UAV targets \cite{2016TPAMIdetecting,2022TIMFang,2023CVPRWLyu,2023TIIFang,2023ACMMMFang}. Infrared UAV targets are challenging due to their weak imaging features and complex backgrounds \cite{2024TPAMIHuang,2024TMJiang,2024ICRAYuan}. TAD \cite{2023CVPRWLyu} leveraged the inconsistent motion cues between UAV targets and the background to detect potential targets. 
Fang et al. \cite{2022TIMFang} formulated the UAV detection task as a residual image prediction by learning the mapping from input images to residual images. 
DAGNet \cite{2023TIIFang} introduced attention mechanisms to adaptively enhance the network's ability to discriminate between UAVs and the background. DANet \cite{2023ACMMMFang} constructed a dynamic attention network for UAVs to enhance feature extraction capabilities. However, the above methods primarily focus on designing complex network structures for improving detection performance, while rarely exploring how to better enhance feature representation without increasing computational complexity. We focus on introducing new auxiliary losses into the backbone to boost the model's ability to represent UAV target features while suppressing the background.



\subsection{Joint Low-Level Image Processing and High-Level Vision Tasks}
Recently, joint approaches for image enhancement and object detection have emerged to further improve detection performance on low-quality images. One category of these methods utilized encoder-decoder architectures for image enhancement \cite{2022AAAILi}, but these modules are complex and hinder real-time performance. Another approach uses classical mathematical models for enhancement, replacing manually designed parameters with predictions from deep networks. IA-YOLO \cite{2022AAAILiu} integrated an adaptive enhancement strategy with YOLOv3 \cite{2018yolov3Redmon} via a differentiable image processing module, improving detection in foggy conditions. BAD-Net \cite{2023TPAMILi} introduced a dual-branch structure to minimize the impact of poor dehazing performance on the detection module. 
However, the above methods are developed for visible light images degraded by adverse weather conditions. To the best of our knowledge, no work has explored handling both NUC and UAV target detection for infrared imaging within a single framework. We propose an end-to-end framework to simultaneously improve both correction and detection performance. 

\section{Detection-Friendly Union Method} \label{sec:method}
\subsection{Overall Architecture}

In this section, we propose a novel network architecture, UniCD, as shown in \cref{fig2}. The UniCD leverages a lightweight prediction network to estimate the bias field and then passes the corrected image to a UAV detection network tailored for UAV targets. 
Additionally, during the joint training of image correction and detection tasks, we employ a detection-guided self-supervised loss to minimize feature discrepancies between the two tasks. Finally, we construct a new dataset, IRBFD, to validate our approach.

\subsection{Prior- and Data-Driven Nonuniformity Correction} \label{subsec:NUCM}
Infrared UAV images captured by thermal detectors often suffer from low-frequency nonuniformity, which significantly impacts target detection performance. Existing model-driven correction methods often struggle to handle complex non-uniformities, while deep learning-based methods suffer from high network complexity. To address these issues, we propose a lightweight correction network that combines parametric prior knowledge with the strong learning capabilities of deep neural networks. The network architecture is shown in \cref{fig2}.

Generally, the degraded infrared image can be represented as follows \cite{2019GRSLChang, 2020TGRSLiu}: \vspace{-.7em} 
\begin{equation}\label{eq1}\vspace{-.7em} 
 Y = C + B, 
\end{equation}
where $Y$, $C$, and $B$ represent the degraded image, the clear image, and the bias field, respectively. Thus, once $B$ is available, the corrected image $R$ can be obtained as $Y-B$. Under ideal conditions, $R$ is theoretically equivalent to the clear image $C$. The bias field $B$ possesses a spatially smooth property, allowing us to model it using the following bivariate polynomial:\vspace{-.5em}  
\begin{equation}\label{eq2}  \vspace{-.5em} 
B\left(x_i,y_j\right) = \sum_{t=0}^{D}\sum_{s=0}^{D-t}a_{t,s}x_i^ty_j^s = \textbf{m}^\top{\textbf{a}},
\end{equation}
where $\left(x_i, y_j\right)$, $D$, and ${\textbf{m}}$ denote the image coordinates, the degree of the polynomial, and the column vector holding the monomial terms, respectively. The column vector ${\textbf{a}}$ represents the coefficients of the polynomial formed by concatenating $\{a_{t,s}\}$. To reduce the redundancy of the basic plane and the computational complexity of higher-order models, we set the degree $D$ to 3.

Accurate estimation of the polynomial coefficients $\textbf{a}$ is crucial for improving bias field correction performance. In this work, we design a lightweight bias field prediction network that can estimate the model parameters accurately and efficiently. 

In our NUC module, we first downsample the degraded image $Y$ by a factor of two, resulting in the downsampled image \( Y_{\text{down}} \). Then, we utilize the global bias field encoder (GBFE) and the local bias field encoder (LBFE) to extract features at different granularities. The GBFE, inspired by the RSTB \cite{2021ICCVWLiang} module, includes two Swin Transformer layers with the hidden layer channel dimension reduced to 16. The LBFE consists of two spatial attention modules in series, allowing it to capture localized features more effectively. The global and local features, \( F_{\text{global}} \) and \( F_{\text{local}} \), are fused to form the final feature representation \( F_{\text{fused}} \):\vspace{-.5em} 
\begin{equation}\label{eq3} \vspace{-.5em} 
F_{\text{fused}} = \text{GBFE}(Y_{\text{down}}) + \text{LBFE}(Y_{\text{down}}).
\end{equation}

Finally, the fused features \( F_{\text{fused}} \) pass through five $3\times3$ convolutional layers ($\text{Conv}_{5}$) and a fully connected (FC) layer to predict the final coefficient vector $\hat{\textbf{a}}$: \vspace{-.5em} 
\begin{equation}\label{eq4} \vspace{-.5em} 
\hat{\textbf{a}}= \text{FC}(\text{Conv}_{5}(F_{\text{fused}})).
\end{equation}

This approach enables the network to adaptively adjust to inputs with varying levels of degradation, effectively predicting the coefficient vector $\hat{\textbf{a}}$  for bias field correction. 

\textbf{Analysis.} Compared to existing model-driven methods, our approach avoids reliance on hand-crafted features. Compared to existing data-driven methods, our approach transforms the high-dimensional image space prediction into a low-dimensional data-driven problem with a few hyperparameters, effectively reducing computational complexity and improving correction performance.

\textbf{Loss Function.} 
We calculate the mean absolute error (MAE) loss between the predicted coefficients $\hat{\textbf{a}}$ and the predefined coefficients $\textbf{a}$ to minimize the discrepancy between them. The MAE loss is defined as:\vspace{-.5em} 
\begin{equation}\vspace{-.5em} 
L_{cor} = \frac{1}{N} \| \mathbf{\hat{a}} - \mathbf{a} \|_2^2,
\label{eq5}
\end{equation}
where $N$ is the total number of elements in the vector $\textbf{a}$. By minimizing this loss, we improve the accuracy of the corrected image, making it closer to the ideal clear image. This loss function is used to pre-train the NUC module separately, serving as the initial weights for the joint training of correction and detection.

\subsection{Mask-Supervised Infrared UAV Detector}\label{subsec:Detect}
\begin{figure}
\centering 
\includegraphics[width=3.2in,keepaspectratio]{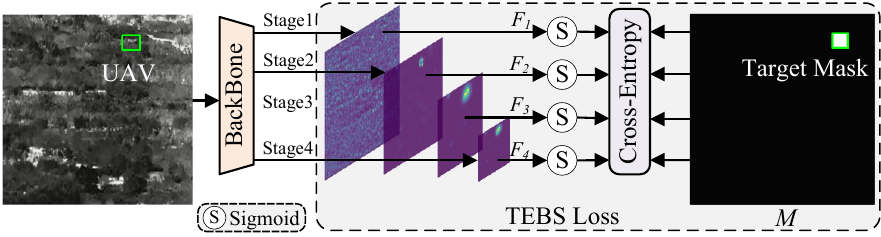}\vspace{-0.6em} 
\caption{Calculation of the proposed feature enhancement and background suppression (TEBS) loss.} \vspace{-1.4em} 
\label{fig3} 
\end{figure}

We select DANet \cite{2023ACMMMFang} as our baseline detector, which constructs a multi-scale dynamic perception network to address the challenges of multi-scale variations in UAV targets. Infrared UAV targets typically have weak features and complex backgrounds, and existing methods often enhance feature extraction through complex model architectures. In this work, we design an auxiliary loss function for further target enhancement and background suppression (TEBS) without increasing network complexity, as illustrated in \cref{fig2,fig3}.

Specifically, we first convert the bounding boxes of the ground-truth into a binary mask $M$, assigning a value of 1 to the target region and 0 to the background region:\vspace{-.7em} 
\begin{equation}\vspace{-.7em} 
M(x, y) = 
\begin{cases} 
   1, & \text{if } (x, y) \in \text{target region}, \\ 
   0, & \text{if } (x, y) \in \text{background region}.
\end{cases}
\end{equation}
This mask is then used to compute a binary cross-entropy loss with the feature maps $ F_i$  from the $i$-th stage of the backbone network. The TEBS loss $ L_{\text{TEBS}}$ is obtained by summing the losses from four stages of the backbone and averaging: \vspace{-.7em}  
\begin{equation}\vspace{-.7em} 
L_{TEBS} = \frac{1}{4} \sum_{i=1}^{4} L_{CE}(M, F_i),  
\end{equation}
where $L_{CE}(\cdot,\cdot)$ denotes the cross-entropy loss. 

The TEBS loss offers three key benefits: (1) Supervision on the target regions helps the backbone network to quickly focus on the feature learning of infrared UAV targets and enhance localization accuracy; (2) Supervision on background regions effectively suppresses non-target features, thereby reducing clutter and noise interference in the features; (3) The loss enhances training efficiency by guiding the network to learn target features more accurately, leading to faster convergence.

The classification and regression losses are kept consistent with the baseline method. The final detector loss is written as:\vspace{-.5em} 
\begin{equation} \vspace{-.5em} 
L_{det} = L_{cls} + L_{reg} + \lambda L_{TEBS},
\label{eq8}
\end{equation}
where $\lambda $ is set to 1 for the first 20 epochs to accelerate convergence and improve localization performance in the early stages. After 20 epochs,  $\lambda $ is reduced to 0.01 to balance with the gradually decreasing loss and avoid impacting classification accuracy.

\subsection{Balance Correction and Detection with Bias-Robust Loss} \label{subsec:Loss}
\begin{figure}
\centering 
\includegraphics[width=3.2in,keepaspectratio]{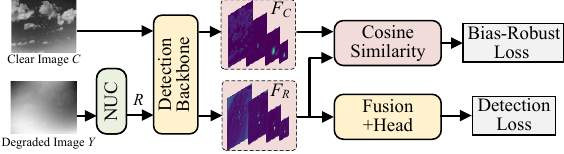}\vspace{-0.7em}  
\caption{Construction of the bias-robust loss in the unified NUC and detection framework.} \vspace{-1.6em}
\label{fig4} 
\end{figure}
Existing research \cite{2022AAAILiu,2023TPAMILi} shows that incorporating supervision losses from low-level vision tasks in joint training can hinder high-level vision task performance. This is because low-level tasks focus on preserving fine image details, while high-level tasks aim to extract target-specific features and ignore irrelevant background information. This conflict can hinder convergence during joint training, causing the high-level vision task to settle into a local optimum.

To address this issue, we design a self-supervised loss, named bias robust (BR) loss, to achieve detection-friendly NUC, as illustrated in \cref{fig4}. 
Specifically, during joint training, the clear image $C$ and the corrected image $R$ obtained via the NUC module are simultaneously fed into the backbone of the detection model. 
Let \( F_{C}^{(i)} \) and \( F_{R}^{(i)} \) represent the feature maps from the \( i \)-th stage of the detection backbone, where \( i = 1, 2, 3, 4 \) denotes the four different stages. Here, clear images are employed only for the joint training of correction and detection on the synthetic dataset and are not applied to NUC training on real infrared images. 

To evaluate the alignment of the corrected image with the clear image in the feature space, we compute the cosine similarity between the feature maps at each stage $i$ as follows: \vspace{-.5em} 
\begin{equation}\vspace{-.5em}   
\text{Cos\_Sim}(F_{C}^{(i)},F_{R}^{(i)}) = \frac{F_{C}^{(i)} \cdot F_{R}^{(i)}}{\left\|F_{C}^{(i)}\right\| \left\|F_{R}^{(i)}\right\|}.
\end{equation}
The details of the cosine similarity calculation are placed in the supplementary materials. The BR loss is obtained by summing the cosine similarities across all stages and averaging them:\vspace{-.5em} 
\begin{equation}\vspace{-.5em} 
L_{BR}= \frac{1}{4}\sum_{i=1}^{4} \left(1 - \text{Cos\_Sim}(F_{C}^{(i)},F_{R}^{(i)})\right).
\end{equation}

The BR loss function is designed to maximize the consistency between the feature representations of the clear and corrected images, thereby enhancing the fidelity of the corrected image within the feature space and ensuring that it retains essential characteristics beneficial for detection. The final union loss function for the joint training process $L_{uni}$ is written as:\vspace{-.5em} 
\begin{equation}
L_{uni}= L_{det} + L_{BR}.
\label{eq11}
\end{equation}

As illustrated in \cref{fig1}, the union loss $L_{uni}$ ensures effective bias field correction while maximizing detection performance by backpropagating through each part of the network. Compared to separate training of NUC and detection networks, this integrated approach enables a more balanced optimization, leading to enhanced results in both correction and detection tasks.

\subsection{IRBFD Dataset}\label{subsec:IRBFD}
We construct a new benchmark called IRBFD, comprising 30,000 synthetic nonuniformity infrared UAV images (IRBFD-syn) and 20,000 real-world infrared UAV images with nonuniformity field (IRBFD-real). The IRBFD-syn subset provides paired degraded and clear images, with the synthesis process based on \cref{eq1,eq2}. The characteristics of the dataset are summarized as follows. (1) Multiple background types. The IRBFD includes multiple complex scenes, such as dense clouds, buildings, forests, urban areas, and sea. (2) Multi-scale variations. The distance between the UAVs and the sensor ranges from 50 meters to 2 kilometers, resulting in multi-scale variations of the targets. (3) Multiple UAV types. Such as the DJI Inspire, Matrice, Phantom, Mavic, and Mini series. All UAV positions are manually annotated. IRBFD serves as a comprehensive resource for evaluating the impact of non-uniformities on UAV target detection in real-world environments. All images have a size of $640\times512$. Additional details can be found in the supplementary material. 

\section{Experiment}
\subsection{Datasets and Evaluation Metrics}
\textbf{Datasets.} 
We use two parts of the IRBFD dataset: IRBFD-syn and IRBFD-real. IRBFD-syn consists of simulated nonuniform infrared images, allowing us to train the model with controlled background images and varying non-uniformities. Training, validation, and testing sets are split in an 8:1:1 ratio. We train on the simulated dataset and directly validate on real-world dataset to demonstrate the generalizability of our method.

\textbf{Evaluation Metrics.} For NUC, we use peak signal-to-noise ratio (PSNR) and structural similarity index (SSIM) as objective evaluation metrics. For target detection, we evaluate the detection performance using precision (P) and recall (R). Lastly, for real-time performance, we use frames per second (FPS) as the evaluation metric. The signal-to-clutter ratio gain (SCRG) is the ratio of SCR in the corrected image to that in the original image, used to evaluate the improvement in target detectability achieved by the correction method.

\subsection{Implementation Details}
We use Adam as the optimizer, with a learning rate of 0.001. The training lasts for 50 epochs, with a weight decay of $10^{-4}$ and a batch size of 4. During training, we only apply random horizontal flipping for data augmentation. Our experiments are conducted on an NVIDIA RTX 4090 with CUDA 12.4 and PyTorch 1.7.
For NUC, we select Liu \cite{2016IPTLiu}, Shi \cite{2022AOShi}, and AHBC \cite{2024TGRSXie} as the traditional bias field correction methods; DMRN \cite{2019GRSLChang} and TV-DIP \cite{2023IPTLiu} as the DL-based correction methods. They are all designed for infrared images. For target detection, we select YOLO11L \cite{2024YOLO11Jocher} (large version), DAGNet \cite{2023TIIFang}, LESPS \cite{2023CVPRYing}, and MSHNet \cite{2024CVPRLiu} as the convolutional neural network (CNN)-based target detection methods. DAGNet, LESPS, and MSHNet are designed for infrared target detection. Deformable DETR \cite{2021ICLRZhu} and DINO \cite{2023ICLRZhang} are the representative Transformer-based detection methods. 

\subsection{Quantitative Results}
\begin{table}[t]
\centering
\setlength\tabcolsep{1.5pt} 
\fontsize{7}{9}\selectfont  
\setlength{\abovecaptionskip}{3pt} 
\caption{Quantitative comparison of the proposed method with SOTA methods on the synthetic dataset IRBFD-syn. {\textbf{Bold}} and \underline{underline} indicate the best and the second best results, respectively.}
\begin{tabular}{c|ccc|ccccccc} 
\hline 
\multirow{2}{*}{Strategy} & \multicolumn{3}{c|}{Module\vspace{-.3em} }  & \multicolumn{5}{c}{Metrics} \\ \cline{2-4}  \cline{5-9}
& NUC & Detection & Pub'Year  & PSNR & SSIM & P & R & FPS \\  \hline
\multirow{6}{*}{Direct} & \multirow{6}{*}{-} & Deformable DETR & ICLR'21  & \multirow{6}{*}{-} & \multirow{6}{*}{-} & 0.614 & 0.630 & 24 \\
 &  & DINO & ICLR'23   &  &  & 0.904 & \underline{0.640} & 26 \\
 &  & DAGNet & TII'23   &  &  & \underline{0.994} & 0.635 & \textbf{43} \\
 &  & LESPS & CVPR'23   &  &  & 0.033 & 0.446 & 12 \\
 &  & MSHNet & CVPR'24   &  &  & 0.407 & 0.421 & 41 \\ 
 &  & YOLO11L & 2024   &  &  & 0.963 &0.602 & \underline{42} \\   \hline
\multirow{10}{*}{Separate} & \multirow{2}{*}{Liu} & YOLO11L & \multirow{2}{*}{IPT'16} & \multirow{2}{*}{16.800} & \multirow{2}{*}{0.8289} & 0.898 & 0.574 & \textless1 \\
 &  & DAGNet   &  &  &  & 0.978 & 0.578 & \textless1 \\ \cline{2-9}
 & \multirow{2}{*}{DMRN} & YOLO11L & \multirow{2}{*}{GRSL'19}  & \multirow{2}{*}{\underline{24.467}} & \multirow{2}{*}{\underline{0.8600}} & 0.923 & 0.550 & 35 \\
 &  & DAGNet   &  &  &  & 0.966 & 0.595 & 36 \\  \cline{2-9}
 & \multirow{2}{*}{Shi} & YOLO11L & \multirow{2}{*}{AO'22}  & \multirow{2}{*}{13.974} & \multirow{2}{*}{0.7783} & 0.924 & 0.455 & \textless1 \\
 &  & DAGNet   &  &  &  & 0.966 & 0.472 & \textless1 \\  \cline{2-9}
 & \multirow{2}{*}{TV-DIP} & YOLO11L & \multirow{2}{*}{IPT'23}   & \multirow{2}{*}{13.397} & \multirow{2}{*}{0.6374} & 0.131 & 0.020 & 29 \\
 &  & DAGNet   &  &  &  & 0.599 & 0.020 & 30 \\  \cline{2-9}
 & \multirow{2}{*}{AHBC} & YOLO11L & \multirow{2}{*}{TGRS'24}   & \multirow{2}{*}{13.954} & \multirow{2}{*}{0.6763} & 0.825 & 0.080 & \textless1 \\
 &  & DAGNet   &  &  &  & 0.724 & 0.040 & \textless1 \\  \hline
Union & \multicolumn{2}{c}{UniCD} & -  & \textbf{31.961} & \textbf{0.9827} & \textbf{0.999} & \textbf{0.822} & 32 \\ \hline
\end{tabular} 
\label{tab:irbfd_simulation}
\end{table}

\begin{table}[t]
\centering
\setlength\tabcolsep{2pt} 
\fontsize{7}{9}\selectfont  
\setlength{\abovecaptionskip}{4.5pt} 
\caption{Quantitative comparison of the proposed method with SOTA methods on the real dataset IRBFD-real. }
\begin{tabular}{ccccccc} \hline
Strategy & NUC & Detection & SCRG & P & R \\   \hline
\multirow{3}{*}{Direct} & \multirow{3}{*}{-} & DINO & \multirow{3}{*}{-} & 0.971 & 0.660 \\
 &  & YOLO11L &  & 0.966 & 0.843 \\
 &  & DAGNet &  & \underline{0.992} & \underline{0.871} \\   \hline
\multirow{4}{*}{Separate} &  \multirow{2}{*}{TV-DIP} & YOLO11L & \multirow{2}{*}{0.412} & 0.521 & 0.024 \\
 &  & DAGNet &  & 0.663 & 0.026 \\  \cline{2-6}
 &  \multirow{2}{*}{AHBC} & YOLO11L & \multirow{2}{*}{\underline{1.146}} & 0.940 & 0.633 \\
 &  & DAGNet &  & {0.986}& 0.699 \\  \hline
Union & \multicolumn{2}{c}{UniCD} & \textbf{1.286} & \textbf{0.994} & \textbf{0.901} \\  \hline
\end{tabular}
\label{tab:irbfd_real} 
\end{table}

\begin{figure}\label{fig5}
\centering 
\includegraphics[width=2.3in,keepaspectratio]{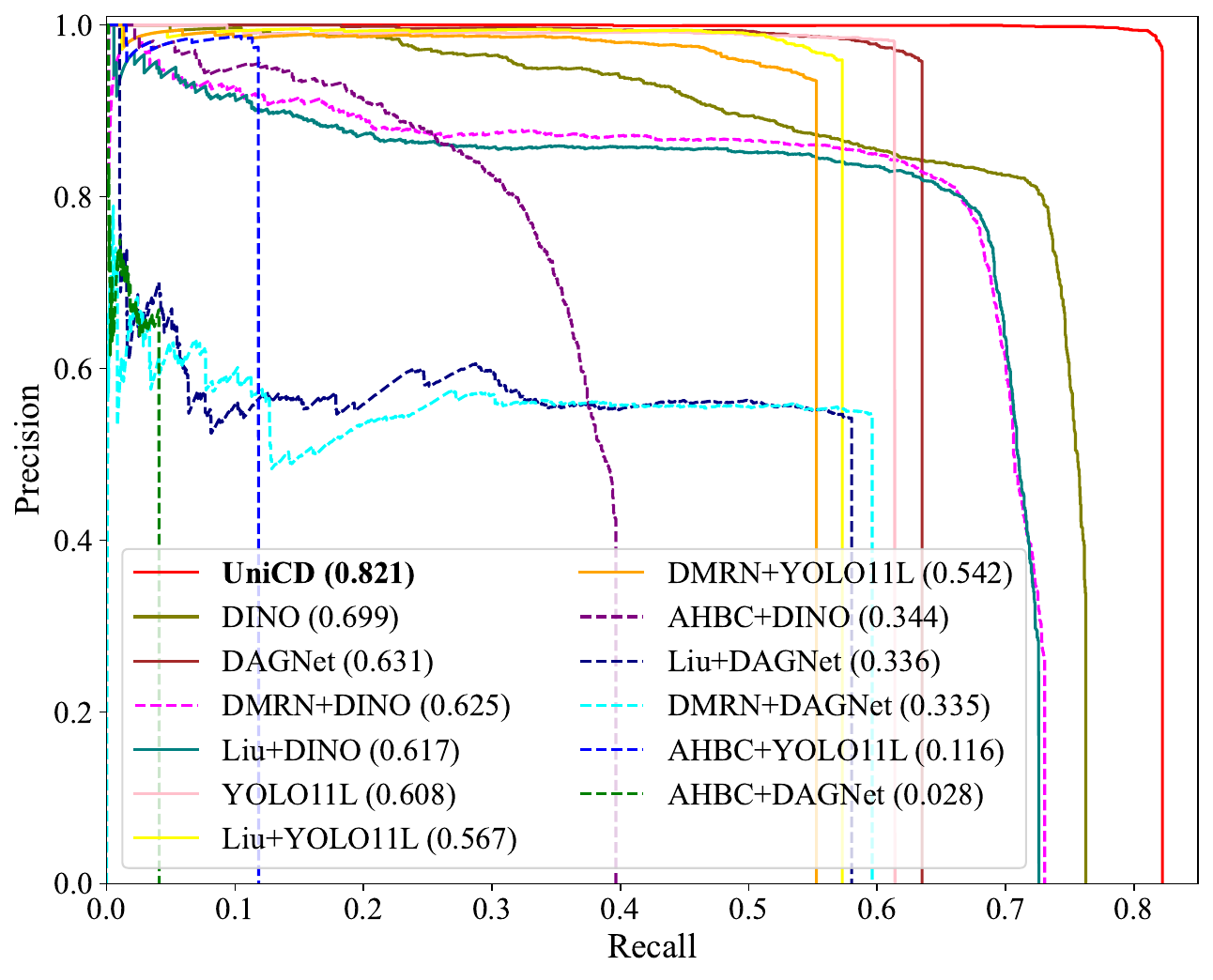}
\setlength{\abovecaptionskip}{6.2pt} 
\vspace{-0.7em}
\caption{P-R curves of our UniCD and other correction-then-detection methods on the IRBFD-syn. The area values under the curves are placed after the method names.} \vspace{-1.5em}
\label{fig:PR_curve} 
\end{figure}

As shown in \cref{tab:irbfd_simulation}, the existing detection methods obtain low P and R values when detecting directly on the degraded bias field images. DAGNet has a high P value of 0.994 but a low recall. This indicates that the bias field has an adverse effect on the UAV target detection. For the separate strategy, the DL-based method DMRN enhances image quality, achieving notable improvements in PSNR and SSIM. Other correction methods, such as Liu, Shi, and AHBC, are less effective for severely degraded images, which further impacts detection accuracy. \Cref{tab:irbfd_real} shows that the three detection methods achieve high P and R when directly detecting on real images with low degradation levels. YOLO11L and DAGNet have lower P and R values for images corrected by TV-DIP, as TV-DIP not only lacks corrective effects but also deteriorates the image content. In contrast, the proposed UniCD achieves the best performance in terms of all the evaluation metrics compared with SOTA methods in \cref{tab:irbfd_simulation} for the synthetic dataset and \cref{tab:irbfd_real}  for the real dataset. Specially, our UniCD achieves a real-time processing speed of 32 FPS. We also plot the R-R curves for our UniCD and other correction-then-detection methods on the IRBFD dataset shown in \cref{fig:PR_curve}. Higher values of the area under the curve indicate better performance. It can be seen that our UniCD achieves the largest area under the curve among all correction-then-detection methods.

\subsection{Qualitative Results}

\begin{figure*}
\centering 
\includegraphics[width=6.7in,keepaspectratio]{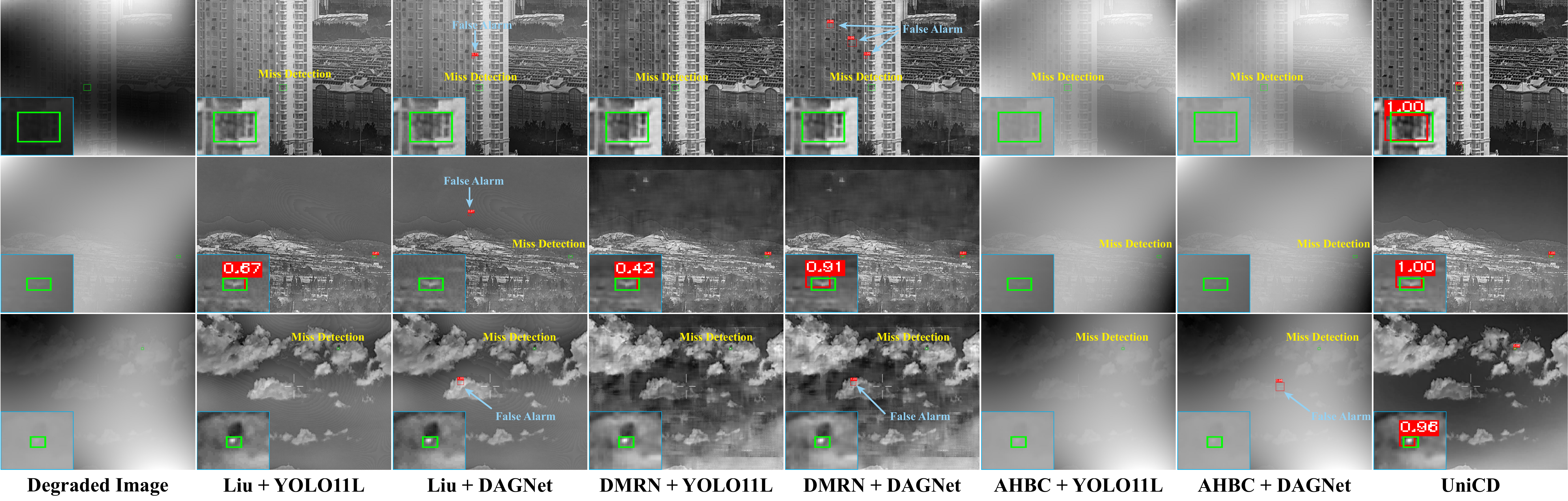} \vspace{-0.8em}
\caption{Visual comparison of results from separate correction followed by detection methods and our UniCD on the IRBFD-syn dataset. Closed-up views are shown in the left bottom corner. Boxes in green and red represent ground-truth and correctly detected targets, respectively.  }\vspace{-1em} 
\label{fig:visual_result_simu} 
\end{figure*}
 
\begin{figure*}
\centering 
\includegraphics[width=6.5in,keepaspectratio]{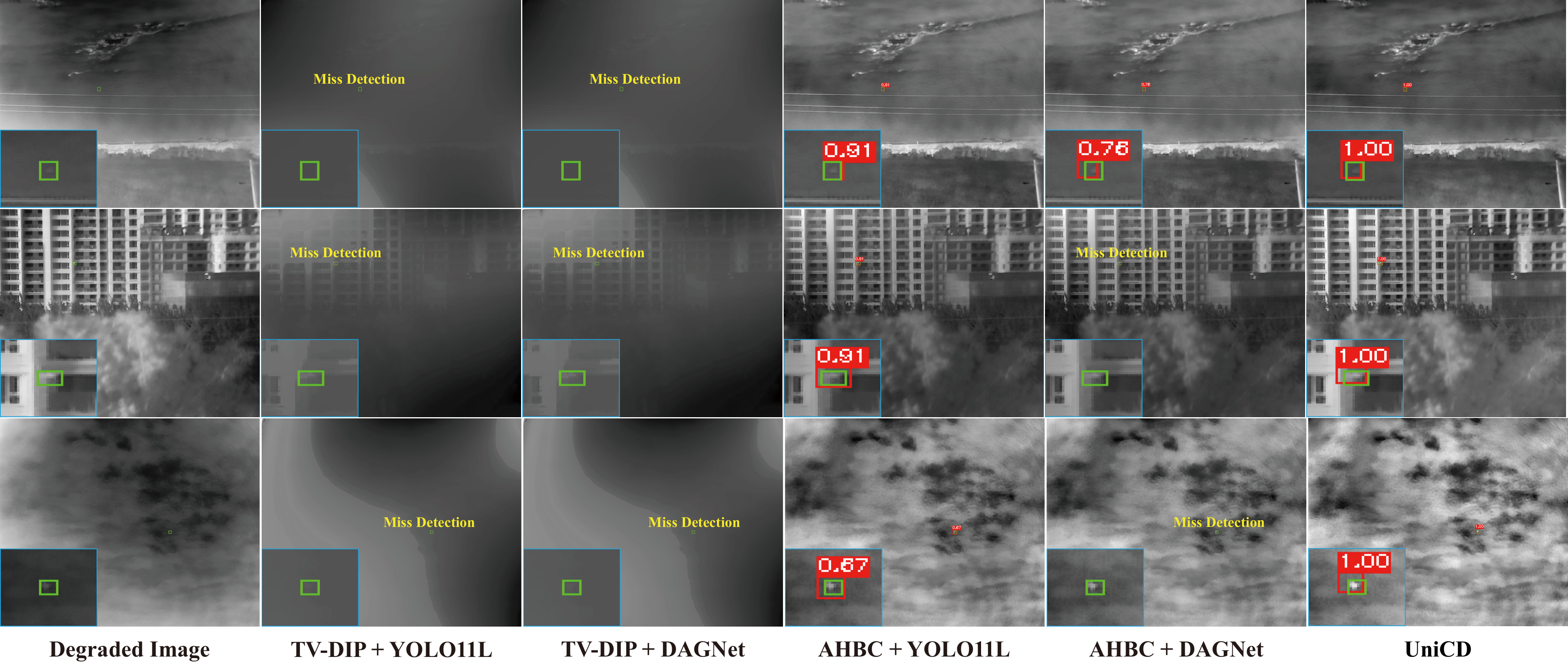} \vspace{-0.7em}
\caption{Visual comparison of results from separate correction-then-detection methods and our UniCD on the IRBFD-real dataset. }\vspace{-1.2em} 
\label{fig:visual_result_real} 
\end{figure*}

As demonstrated in \cref{fig:visual_result_simu}, we show qualitative results from various separate correction-then-detection methods and our UniCD on the IRBFD-syn dataset across three distinct scenarios: buildings, hillside, and clouds. As can be seen, even in severe degraded bias field situations, our UniCD can still perform high-quality image correction while accurately detecting the UAV targets. This is because the proposed NUC correction module integrates parametric modeling and a small number of model parameter predictions, enabling more accurate parameter estimation. Meanwhile, our detection network introduces auxiliary loss with target mask supervision into the backbone to enhance the features of UAV targets while suppressing the background, thereby improving detection performance. Conventional correction methods, such as Liu and AHBC, have limited modeling capabilities and are prone to producing bias field residuals. The corrected image from DMRN exhibits block artifacts. Existing detection methods produce false alarms or miss detections on the above-corrected images; similar results are observed in \cref{fig:visual_result_real} for scenes degraded by real bias fields. See the supplementary material for more visual results.

\subsection{Ablation Study}
In this section, we report ablation study results.
\begin{table}[t]
\centering
\setlength\tabcolsep{2pt} 
\fontsize{7}{9}\selectfont  
\setlength{\abovecaptionskip}{3pt} 
\caption{Ablation study of polynomial degrees.} 
\begin{tabular}{ccccccc} \hline
Degree & Number of coefficients& PSNR & SSIM  & P & R &\\  \hline
2 & 6 & 13.5279 & 0.7590 & 0.991 & 0.433 & \\
3 &  10 &\textbf{39.050} & \textbf{0.9970} &  \textbf{0.997} &  \textbf{0.810}  \\ 
4 &  15 & 31.744 & 0.9890 & \textbf{0.997} & 0.788\\ 
5 &  21 &29.070  & 0.9830 & \textbf{0.997}& 0.787 \\  \hline
\end{tabular} 
\label{tab:degree_selection}\vspace{-1em} 
\end{table}
{\textbf{Impact of polynomial degrees.}} We conduct experiments to determine the optimal polynomial degree, as described in \cref{tab:degree_selection}. The results indicate that the highest values for all correction and detection metrics are achieved when the degree is set to 3. Compared to higher-order polynomials, a third-order polynomial has lower complexity and less redundance. A lower degree implies weaker modeling capability, resulting in poor correction performance.

\begin{table}[t]
\centering
\setlength\tabcolsep{2pt}
\setlength{\abovecaptionskip}{3pt} 
\caption{Ablation study of the LBFE and GBFE modules.}
\fontsize{8}{9}\selectfont  
\begin{tabular}{ccccccc}  \hline
LBFE & GBFE & Params (M) & FLOPs (G) & PSNR & SSIM & FPS\\  \hline
$\times$ & $\times$ & 0.3786 & 0.1192 &  27.218 & 0.9808&555 \\ 
$\surd$ &  $\times$   & 0.3812 & 0.2747 & 30.242 & 0.9877&370 \\
$\times$ & $\surd$ & 0.3940 & 1.5159 & 37.888 & 0.9961& 151\\
$\surd$ & $\surd$ &  0.3966 & 1.6809 &\textbf{39.050} & \textbf{0.9970}& 116\\   \hline
\end{tabular}
\label{tab:twobranch}\vspace{-1em}
\end{table}
 
{\textbf{Impact of the LBFE and GBFE components.}} As presented in \cref{tab:twobranch}, when using only LBFE or GBFE, we can see that the PSNR of the correction model already achieves 30.242 and 37.888, respectively, surpassing the 27.218 of the baseline. When combined with LBFE and GBFE, further improvements can be achieved, reaching 39.050 with only 1.6809G FLOPs and 0.3966M parameters on an image size of $640\times512$. Additionally, the SSIM value has also shown some slight improvements. This suggests that the Transformer architecture combined with the CNN structure can effectively model bias fields, significantly improving performance while maintaining high real-time efficiency. 

\begin{table}[t]
\centering
\fontsize{7}{9}\selectfont  
\setlength{\abovecaptionskip}{3pt} 
\caption{Ablation study of the auxiliary TEBS loss.}
\setlength{\belowcaptionskip}{-0.2cm}
\begin{tabular}{ccccccc} \hline
TEBS Loss & P & R \\   \hline
w/o & 0.993 & 0.762  \\  \hline
w & \textbf{0.997} & \textbf{0.810} \\   \hline
\end{tabular}
\label{tab:auxiliaryloss}
\end{table}

\begin{figure}
\centering 
\includegraphics[width=2.8in,keepaspectratio]{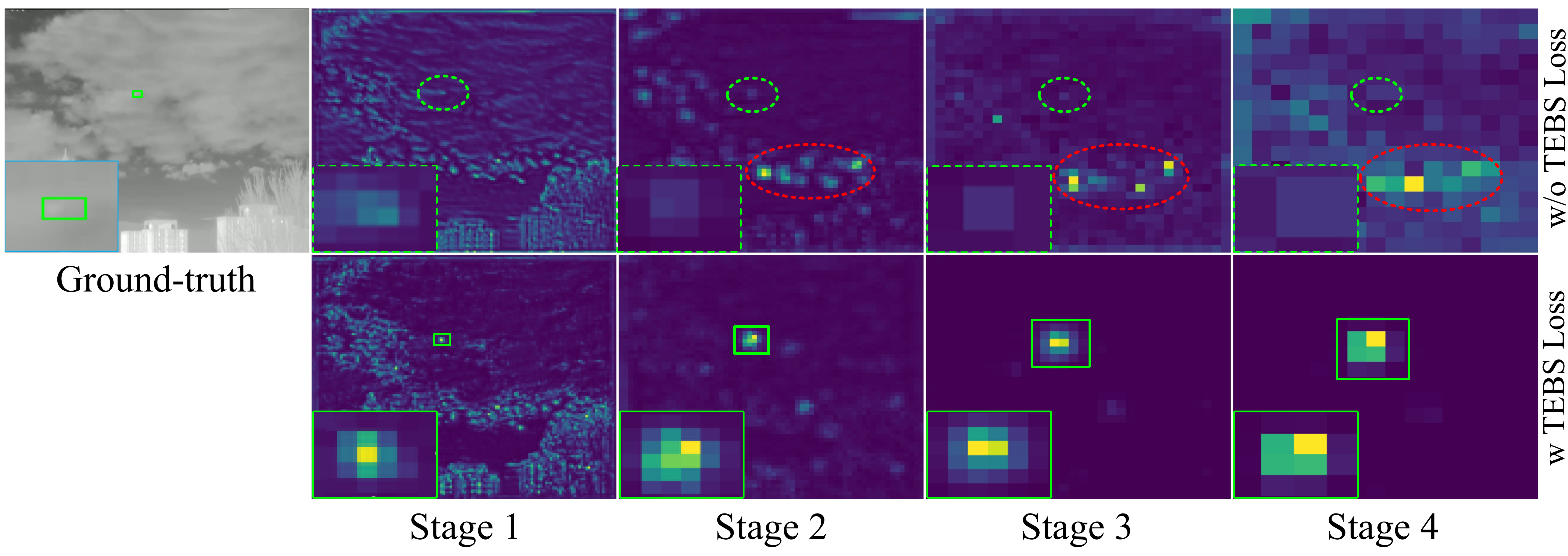} \vspace{-0.6em} 
\caption{Comparison of feature maps from the detection backbone at different stages with and without TEBS loss. }\vspace{-0.5em} 
\label{fig:visual_result_TEBSloss} 
\end{figure}
{\textbf{Impact of the auxiliary TEBS loss.}} 
As shown in \cref{tab:auxiliaryloss}, the auxiliary TEBS loss added to the backbone of the UAV detection network leads to improvements in both P and R performance. The TEBS loss imposes strong constraints on the target and background masks for the backbone features, enhancing target features while suppressing the background, thus improving detection performance, as depicted in \cref{fig:visual_result_TEBSloss}. 
In the absence of TEBS loss, the targets are generally weak and there is considerable residual background.

\begin{table}[t]
\centering
\setlength\tabcolsep{2pt} 
\fontsize{7}{9.5}\selectfont  
\setlength{\abovecaptionskip}{1pt} 
\caption{Ablation study of BR loss on the synthetic dataset IRBFD-syn. Here, the direct and separate strategies utilize the correction and detection modules proposed in UniCD.}
\begin{tabular}{cccccccc} \hline
\multirow{2}{*}{Strategy} &\multirow{2}{*}{ \makecell{Correction\\loss}}&\multirow{2}{*}{\makecell{Detection\\loss}}& \multirow{2}{*}{\makecell{BR\\loss}}& \multicolumn{4}{c}{Metrics} \\  \cline{5-8}
 &&&& PSNR & SSIM & P & R \\ \hline
Direct & $\times$& $\surd$ & $\times$  & - & - & 0.998 & 0.694 \\
 Separate & $\surd$& $\surd$ & $\times$ & \textbf{37.722} & \textbf{0.9960} & \textbf{0.999} & {0.793} \\ 
 Union & $\surd$& $\surd$ & $\times$  & 33.024 & 0.9827  &0.998   & 0.791 \\
Union & $\times$& $\surd$ & $\times$ & 17.940 & 0.8910 &  0.989& 0.811\\
UniCD & $\times$& $\surd$ & $\surd$ & 31.961 & 0.9827 & \textbf{0.999} & \textbf{0.822} \\  \hline
\end{tabular}
\label{tab:framework_ablation_syn}
\end{table}

{\textbf{Impact of BR loss.}} As shown in \cref{tab:framework_ablation_syn}, the direct detection results in low P and R. The correction-then-detection separate method without BR loss in the second row obtains high PSNR, SSIM and P, but with low R because of the independent processing of the two tasks. The union of correction and detection without BR loss in the third row leads to a decrease in PSNR, SSIM, P, and R due to the conflict between the two tasks. The union of correction and detection without correction and BR losses in the fourth row significantly reduces PSNR and SSIM owing to the detection module's sole constraint. Our UniCD achieves the highest R value and P value exceeding 0.99, demonstrating that our union framework effectively balances correction and detection through the self-supervised BR loss.

{\textbf{Effectiveness of UniCD on real dataset.}} 
As shown in \cref{tab:framework_ablation_real}, we use the NUC module weights trained on the IRBFD-syn and directly test them on the IRBFD-real. This configuration is referred to as $\text{UniCD}^{\star}$. The direct and separate methods with the modules in UniCD achieve a high P but a low R. Our UniCD obtains the highest R value and P value exceeding 0.99, thereby validating the effectiveness and generalization of our union framework on the real dataset.

{\textbf{Generalization of the NUC Module.}} 
In \cref{tab:k value}, we test the UniCD on images with various degrees of non-uniformity degradation without retraining the NUC module. We control the severity of degradation using the formula $Y=C+k\ast B$, where the $k$-value determines the level of degradation. The correction results at different $k$-values indicate that the NUC module generalizes well to various degradation levels.

\begin{table}[t]
\centering
\begin{minipage}{0.23\textwidth}  
\centering
\setlength\tabcolsep{2.2pt} 
\fontsize{7}{9}\selectfont  
\setlength{\abovecaptionskip}{3.5pt} 
\caption{Ablation study of our UniCD on the IRBFD-real.}
\begin{tabular}{cccc} \hline
 Strategy & SCRG & P & R  \\  \hline
Direct & - & 0.992 & 0.887  \\
Separate & \textbf{1.286} & \textbf{0.998} & {0.812}  \\
$\text{UniCD}^{\star}$& \textbf{1.286} & 0.994 & \textbf{0.901}  \\  \hline
\end{tabular}
\label{tab:framework_ablation_real}
\end{minipage} \hfill
\begin{minipage}{0.23\textwidth}  
\centering
\setlength\tabcolsep{2.2pt} 
\fontsize{7}{9}\selectfont  
\setlength{\abovecaptionskip}{3.5pt} 
\caption{Ablation study of varying levels of nonuniformity.}
\setlength{\belowcaptionskip}{-0.2cm}
\begin{tabular}{ccccccc} \hline
K-value & PSNR & SSIM \\   \hline
3 & 29.119 & 0.9891 \\  \hline
5 & 34.907 & 0.9950 \\  \hline
12 & 38.361 & 0.9968 \\   \hline
\end{tabular}
\label{tab:degree_selection}\vspace{-1em} 
\label{tab:k value}
\end{minipage}
\end{table}

\begin{table}[t]
\centering
\setlength\tabcolsep{2pt} 
\fontsize{7}{9}\selectfont  
\setlength{\abovecaptionskip}{3pt} 
\caption{Ablation study of the union of our scalable NUC module with existing detection methods.}
\begin{tabular}{ccccc} \hline
 Our NUC & Detection & P & R &FPS\\   \hline
\multirow{4}{*}{ $\times$} & YOLO11L & 0.835 & 0.075& 42\\
 & DAGNet & 0.711 & 0.036& 43\\
 & LESPS & 0.007 & 0.122 &12\\
 & MSHNet & 0.284 & 0.221 &42\\  \hline
\multirow{4}{*}{$\surd$} & YOLO11L & $0.977_{(+0.142)}$ & $0.657_{(+0.582)}$& 31\\
 & DAGNet & $0.997_{(+0.286)}$ & $0.722_{(+0.686)}$ &31\\
 & LESPS & $0.007_{(+0.000)}$ & $0.459_{(+0.337)}$& 11\\
 & MSHNet & $0.776_{(+0.492)}$ & $0.701_{(+0.480)}$& 31\\  \hline
\end{tabular}
\setlength{\belowcaptionskip}{3pt} 
\label{tab:NUC_plug_and_play}
\end{table}
{\textbf{Scalability of the NUC Module.}} 
From \cref{tab:NUC_plug_and_play}, we observe that, except for the P value of LESPS, introducing our correction module significantly boosts the detection performance of several recent general and infrared target detection methods. This indicates that our NUC module can be flexibly integrated as a scalable component into existing detectors to enhance infrared images for detection purposes.

\section{Conclusion}  \label{sec:conclusion}
In this paper, we propose UniCD, an end-to-end framework that simultaneously addresses bias field correction and infrared UAV target detection. We develop a NUC module that removes bias fields and restores clear images with parameters adaptively predicted by a lightweight network. Additionally, we introduce auxiliary losses with mask supervision to enhance UAV target features and suppress the background. We also present a self-supervised feature loss to improve the robustness of detection to varying bias levels. Moreover, we construct a new dataset IRBFD to facilitate future research.  Experimental results show that our UniCD outperforms previous approaches in both synthetic and real-world scenarios. Furthermore, our method shows great potential for deployment on resource-constrained edge devices.

\noindent
\textbf{Acknowledgments.} This work was supported by the Open Research Fund of the National Key Laboratory of Multispectral Information Intelligent Processing Technology under Grant 61421132301, and the Natural Science Foundation of Jiangsu Province under Grant BK202302028.

{
    \small
    \bibliographystyle{ieeenat_fullname}
    \bibliography{references}

\begin{thebibliography}{34}
\providecommand{\natexlab}[1]{#1}
\providecommand{\url}[1]{\texttt{#1}}
\expandafter\ifx\csname urlstyle\endcsname\relax
  \providecommand{\doi}[1]{doi: #1}\else
  \providecommand{\doi}{doi: \begingroup \urlstyle{rm}\Url}\fi

\bibitem[Cao and Tisse(2014)]{2014OLCao}
Yanpeng Cao and Christel-Loic Tisse.
\newblock Single-image-based solution for optics temperaturedependent nonuniformity correction in an uncooled long-wave infrared camera.
\newblock \emph{Optics Letters}, 39\penalty0 (3):\penalty0 646--648, 2014.

\bibitem[Chang et~al.(2019)Chang, Yan, Liu, Fang, and Zhong]{2019GRSLChang}
Yi Chang, Luxin Yan, Li Liu, Houzhang Fang, and Sheng Zhong.
\newblock Infrared aerothermal nonuniform correction via deep multiscale residual network.
\newblock \emph{IEEE Geoscience and Remote Sensing Letters}, 16\penalty0 (7):\penalty0 1120 -- 1124, 2019.

\bibitem[Fang et~al.(2022)Fang, Ding, Wang, Chang, Yan, and Han]{2022TIMFang}
Houzhang Fang, Lan Ding, Liming Wang, Yi Chang, Luxin Yan, and Jinhui Han.
\newblock Infrared small {UAV} target detection based on depthwise separable residual dense network and multiscale feature fusion.
\newblock \emph{IEEE Transactions on Instrumentation and Measurement}, 71:\penalty0 1--20, 2022.

\bibitem[Fang et~al.(2023{\natexlab{a}})Fang, Liao, Wang, Li, Chang, Yan, and Wang]{2023ACMMMFang}
Houzhang Fang, Zikai Liao, Lu Wang, Qingshan Li, Yi Chang, Luxin Yan, and Xuhua Wang.
\newblock {DANet}: Multi-scale {UAV} target detection with dynamic feature perception and scale-aware knowledge distillation.
\newblock In \emph{Proceedings of the 31st ACM International Conference on Multimedia}, pages 2121--2130, 2023{\natexlab{a}}.

\bibitem[Fang et~al.(2023{\natexlab{b}})Fang, Liao, Wang, Chang, and Yan]{2023TIIFang}
Houzhang Fang, Zikai Liao, Xuhua Wang, Yi Chang, and Luxin Yan.
\newblock Differentiated attention guided network over hierarchical and aggregated features for intelligent {UAV} surveillance.
\newblock \emph{IEEE Transactions on Industrial Informatics}, 19\penalty0 (9):\penalty0 9909--9920, 2023{\natexlab{b}}.

\bibitem[Gella et~al.(2023)Gella, Zhang, Upadhyay, Chang, Waliman, Ba, Wong, and Kadambi]{2023weatherproof}
Blake Gella, Howard Zhang, Rishi Upadhyay, Tiffany Chang, Matthew Waliman, Yunhao Ba, Alex Wong, and Achuta Kadambi.
\newblock {WeatherProof}: A paired-dataset approach to semantic segmentation in adverse weather.
\newblock \emph{arXiv preprint arXiv:2312.09534}, 2023.

\bibitem[Huang et~al.(2024)Huang, Li, Chen, Wang, Zhao, and Xu]{2024TPAMIHuang}
Bo Huang, Jianan Li, Junjie Chen, Gang Wang, Jian Zhao, and Tingfa Xu.
\newblock {Anti-UAV410}: A thermal infrared benchmark and customized scheme for tracking drones in the wild.
\newblock \emph{IEEE Transactions on Pattern Analysis and Machine Intelligence}, 46\penalty0 (5):\penalty0 2852--2865, 2024.

\bibitem[Jiang et~al.(2023)Jiang, Wang, Peng, Yu, Wang, Xing, Li, Guo, Ye, Jiao, Zhao, and Han]{2024TMJiang}
Nan Jiang, Kuiran Wang, Xiaoke Peng, Xuehui Yu, Qiang Wang, Junliang Xing, Guorong Li, Guodong Guo, Qixiang Ye, Jianbin Jiao, Jian Zhao, and Zhenjun Han.
\newblock {Anti-UAV}: A large-scale benchmark for vision-based {UAV} tracking.
\newblock \emph{IEEE Transactions on Multimedia}, 25:\penalty0 486--500, 2023.

\bibitem[Jocher and Qiu(2024)]{2024YOLO11Jocher}
Glenn Jocher and Jing Qiu.
\newblock {Ultralytics YOLO11}, 2024.

\bibitem[Jocher et~al.(2023)Jocher, Chaurasia, and Qiu]{2023YOLOv8Jocher}
Glenn Jocher, Ayush Chaurasia, and Jing Qiu.
\newblock {Ultralytics YOLOv8}, 2023.

\bibitem[Li et~al.(2023)Li, Zhou, Liu, Yang, Xie, Li, and Zhu]{2023TPAMILi}
Chengyang Li, Heng Zhou, Yang Liu, Caidong Yang, Yongqiang Xie, Zhongbo Li, and Liping Zhu.
\newblock Detection-friendly dehazing: Object detection in real-world hazy scenes.
\newblock \emph{IEEE Transactions on Pattern Analysis and Machine Intelligence}, 45\penalty0 (7):\penalty0 8284--8295, 2023.

\bibitem[Li et~al.(2022)Li, Chang, Yu, and Yan]{2022AAAILi}
Yi Li, Yi Chang, Changfeng Yu, and Luxin Yan.
\newblock Close the loop: A unified bottom-up and top-down paradigm for joint image deraining and segmentation.
\newblock In \emph{Proceedings of the AAAI Conference on Artificial Intelligence}, pages 1438--1446, 2022.

\bibitem[Liang et~al.(2021)Liang, Cao, Sun, Zhang, Van~Gool, and Timofte]{2021ICCVWLiang}
Jingyun Liang, Jiezhang Cao, Guolei Sun, Kai Zhang, Luc Van~Gool, and Radu Timofte.
\newblock {SwinIR}: Image restoration using swin transformer.
\newblock In \emph{Proceedings of the IEEE/CVF International Conference on Computer Vision Workshops (ICCVW)}, pages 1833--1844, 2021.

\bibitem[Liu et~al.(2023)Liu, Chen, Bao, and Wang]{2023IPTLiu}
Kang Liu, Honglei Chen, Wenzhong Bao, and Jianlu Wang.
\newblock Thermal imaging spatial noise removal via deep image prior and step-variable total variation regularization.
\newblock \emph{Infrared Physics \& Technology}, 134:\penalty0 104888, 2023.

\bibitem[Liu et~al.(2016)Liu, Yan, Zhao, Dai, and Zhang]{2016IPTLiu}
Li Liu, Luxin Yan, Hui Zhao, Xiaobing Dai, and Tianxu Zhang.
\newblock Correction of aeroheating-induced intensity nonuniformity in infrared images.
\newblock \emph{Infrared Physics and Technology}, 76:\penalty0 235--241, 2016.

\bibitem[Liu et~al.(2020)Liu, Xu, and Fang]{2020TGRSLiu}
Li Liu, Luping Xu, and Houzhang Fang.
\newblock Simultaneous intensity bias estimation and stripe noise removal in infrared images using the global and local sparsity constraints.
\newblock \emph{IEEE Transactions on Geoscience and Remote Sensing}, 58\penalty0 (3):\penalty0 1777 -- 1789, 2020.

\bibitem[Liu et~al.(2024)Liu, Liu, Zheng, Wang, and Fu]{2024CVPRLiu}
Qiankun Liu, Rui Liu, Bolun Zheng, Hongkui Wang, and Ying Fu.
\newblock Infrared small target detection with scale and location sensitivity.
\newblock In \emph{Proceedings of the IEEE/CVF Conference on Computer Vision and Pattern Recognition}, pages 17490--17499, 2024.

\bibitem[Liu et~al.(2022)Liu, Ren, Yu, Guo, Zhu, and Zhang]{2022AAAILiu}
Wenyu Liu, Gaofeng Ren, Runsheng Yu, Shi Guo, Jianke Zhu, and Lei Zhang.
\newblock Image-adaptive {YOLO} for object detection in adverse weather conditions.
\newblock In \emph{Proceedings of the AAAI Conference on Artificial Intelligence}, pages 1792--1800, 2022.

\bibitem[Lyu et~al.(2023)Lyu, Liu, Li, Guo, and Fu]{2023CVPRWLyu}
Yanyi Lyu, Zhunga Liu, Huandong Li, Dongxiu Guo, and Yimin Fu.
\newblock A real-time and lightweight method for tiny airborne object detection.
\newblock In \emph{Proceedings of IEEE/CVF Conference on Computer Vision and Pattern Recognition Workshops (CVPRW)}, pages 3016--3025, 2023.

\bibitem[Redmon and Farhadi(2018)]{2018yolov3Redmon}
Joseph Redmon and Ali Farhadi.
\newblock {YOLOv3}: An incremental improvement, 2018.

\bibitem[Rozantsev et~al.(2017)Rozantsev, Lepetit, and Fua]{2016TPAMIdetecting}
Artem Rozantsev, Vincent Lepetit, and Pascal Fua.
\newblock Detecting flying objects using a single moving camera.
\newblock \emph{IEEE Transactions on Pattern Analysis and Machine Intelligence}, 39\penalty0 (5):\penalty0 879--892, 2017.

\bibitem[Shi et~al.(2022)Shi, Chen, Hong, Zhang, Sang, and Zhang]{2022AOShi}
Yu Shi, Jisong Chen, Hanyu Hong, Yaozong Zhang, Nong Sang, and Tianxu Zhang.
\newblock Multi-scale thermal radiation effects correction via a fast surface fitting with {Chebyshev} polynomials.
\newblock \emph{Applied Optics}, 61\penalty0 (25):\penalty0 7498 -- 7507, 2022.

\bibitem[Shi et~al.(2024)Shi, Zhou, Ma, Wang, and Hong]{2024TGRSShi}
Yu Shi, Yixin Zhou, Lei Ma, Lei Wang, and Hanyu Hong.
\newblock {WDTSNet}: Wavelet decomposition two-stage network for infrared thermal radiation effect correction.
\newblock \emph{IEEE Transactions on Geoscience and Remote Sensing}, 62:\penalty0 1 -- 17, 2024.

\bibitem[VidalMata et~al.(2021)VidalMata, Banerjee, RichardWebster, Albright, Davalos, McCloskey, Miller, Tambo, Ghosh, Nagesh, Yuan, Hu, Wu, Yang, Zhang, Liu, Wang, Chen, Huang, Chin, Li, Lababidi, Otto, and Scheirer]{2020TPAMIBridging}
Rosaura~G. VidalMata, Sreya Banerjee, Brandon RichardWebster, Michael Albright, Pedro Davalos, Scott McCloskey, Ben Miller, Asong Tambo, Sushobhan Ghosh, Sudarshan Nagesh, Ye Yuan, Yueyu Hu, Junru Wu, Wenhan Yang, Xiaoshuai Zhang, Jiaying Liu, Zhangyang Wang, Hwann-Tzong Chen, Tzu-Wei Huang, Wen-Chi Chin, Yi-Chun Li, Mahmoud Lababidi, Charles Otto, and Walter~J. Scheirer.
\newblock Bridging the gap between computational photography and visual recognition.
\newblock \emph{IEEE Transactions on Pattern Analysis and Machine Intelligence}, 43\penalty0 (12):\penalty0 4272--4290, 2021.

\bibitem[Wang et~al.(2024)Wang, Sui, Wang, Liu, Zhang, and Chen]{2024OEWang}
Yu Wang, Xiubao Sui, Yihong Wang, Tong Liu, Chuncheng Zhang, and Qian Chen.
\newblock Contrast enhancement method in aero thermal radiation images based on cyclic multi-scale illumination self-similarity and gradient perception regularization.
\newblock \emph{Optics Express}, 32\penalty0 (2):\penalty0 1650--1668, 2024.

\bibitem[Xie et~al.(2024)Xie, Song, and Huang]{2024TGRSXie}
Jun Xie, Lingfei Song, and Hua Huang.
\newblock Thermal radiation bias correction for infrared images using {Huber} function-based loss.
\newblock \emph{IEEE Transactions on Geoscience and Remote Sensing}, 62:\penalty0 1--15, 2024.

\bibitem[Yang et~al.(2020)Yang, Yuan, Ren, Liu, Scheirer, Wang, Zhang, Zhong, Xie, Pu, Zheng, Qu, Xie, Chen, Li, Hong, Jiang, Yang, Liu, Qu, Wan, Zheng, Zhong, Su, He, Guo, Zhao, Zhu, Liang, Wang, Chen, Quan, Xu, Liu, Liu, Sun, Lin, Li, Lu, Gu, Zhou, Cao, Zhang, Chi, Zhuang, Lei, Li, Wang, Liu, Yi, Zuo, Chi, Wang, Wang, Liu, Gao, Chen, Guo, Li, Zhong, Huang, Guo, Yang, Liao, Yang, Zhou, Feng, and Qin]{2020TIPYang}
Wenhan Yang, Ye Yuan, Wenqi Ren, Jiaying Liu, Walter~J. Scheirer, Zhangyang Wang, Taiheng Zhang, Qiaoyong Zhong, Di Xie, Shiliang Pu, Yuqiang Zheng, Yanyun Qu, Yuhong Xie, Liang Chen, Zhonghao Li, Chen Hong, Hao Jiang, Siyuan Yang, Yan Liu, Xiaochao Qu, Pengfei Wan, Shuai Zheng, Minhui Zhong, Taiyi Su, Lingzhi He, Yandong Guo, Yao Zhao, Zhenfeng Zhu, Jinxiu Liang, Jingwen Wang, Tianyi Chen, Yuhui Quan, Yong Xu, Bo Liu, Xin Liu, Qi Sun, Tingyu Lin, Xiaochuan Li, Feng Lu, Lin Gu, Shengdi Zhou, Cong Cao, Shifeng Zhang, Cheng Chi, Chubing Zhuang, Zhen Lei, Stan~Z. Li, Shizheng Wang, Ruizhe Liu, Dong Yi, Zheming Zuo, Jianning Chi, Huan Wang, Kai Wang, Yixiu Liu, Xingyu Gao, Zhenyu Chen, Chang Guo, Yongzhou Li, Huicai Zhong, Jing Huang, Heng Guo, Jianfei Yang, Wenjuan Liao, Jiangang Yang, Liguo Zhou, Mingyue Feng, and Likun Qin.
\newblock Advancing image understanding in poor visibility environments: A collective benchmark study.
\newblock \emph{IEEE Transactions on Image Processing}, 29:\penalty0 5737--5752, 2020.

\bibitem[Ying et~al.(2023)Ying, Liu, Wang, Li, Chen, Lin, Sheng, and Zhou]{2023CVPRYing}
Xinyi Ying, Li Liu, Yingqian Wang, Ruojing Li, Nuo Chen, Zaiping Lin, Weidong Sheng, and Shilin Zhou.
\newblock Mapping degeneration meets label evolution: Learning infrared small target detection with single point supervision.
\newblock In \emph{Proceedings of the IEEE/CVF Conference on Computer Vision and Pattern Recognition}, pages 15528--15538, 2023.

\bibitem[Yuan et~al.(2024)Yuan, Yang, Nguyen, Nguyen, Yang, Liu, Li, Wang, and Xie]{2024ICRAYuan}
Shenghai Yuan, Yizhuo Yang, Thien~Hoang Nguyen, Thien-Minh Nguyen, Jianfei Yang, Fen Liu, Jianping Li, Han Wang, and Lihua Xie.
\newblock {MMAUD}: A comprehensive multi-modal {Anti-UAV} dataset for modern miniature drone threats.
\newblock In \emph{2024 IEEE International Conference on Robotics and Automation (ICRA)}, pages 2745--2751, 2024.

\bibitem[Zhang et~al.(2023)Zhang, Li, Liu, Zhang, Su, Zhu, Ni, and Shum]{2023ICLRZhang}
Hao Zhang, Feng Li, Shilong Liu, Lei Zhang, Hang Su, Jun Zhu, Lionel Ni, and Heung-Yeung Shum.
\newblock {DINO:} {DETR} with improved denoising anchor boxes for end-to-end object detection.
\newblock In \emph{Proceedings of International Conference on Learning Representations (ICLR)}, 2023.

\bibitem[Zhang et~al.(2022)Zhang, Zhang, Yang, Bai, Zhang, and Guo]{2022CVPRZhang}
Mingjin Zhang, Rui Zhang, Yuxiang Yang, Haichen Bai, Jing Zhang, and Jie Guo.
\newblock {ISNet:} shape matters for infrared small target detection.
\newblock In \emph{Proceedings of IEEE/CVF Conference on Computer Vision and Pattern Recognition (CVPR)}, pages 867--876, 2022.

\bibitem[Zhang et~al.(2024)Zhang, Yang, Guo, Li, Gao, and Zhang]{2024AAAIZhang}
Mingjin Zhang, Handi Yang, Jie Guo, Yunsong Li, Xinbo Gao, and Jing Zhang.
\newblock {IRPruneDet}: efficient infrared small target detection via wavelet structure-regularized soft channel pruning.
\newblock In \emph{Proceedings of the AAAI Conference on Artificial Intelligence}, pages 7224--7232, 2024.

\bibitem[Zheng and Gee(2010)]{2010CVPRZheng}
Yuanjie Zheng and James~C. Gee.
\newblock Estimation of image bias field with sparsity constraints.
\newblock In \emph{Proceedings of the IEEE Conference on Computer Vision and Pattern Recognition}, pages 255--262, 2010.

\bibitem[Zhu et~al.(2021)Zhu, Su, Lu, Li, Wang, and Dai]{2021ICLRZhu}
Xizhou Zhu, Weijie Su, Lewei Lu, Bin Li, Xiaogang Wang, and Jifeng Dai.
\newblock {Deformable DETR}: Deformable transformers for end-to-end object detection.
\newblock In \emph{Proceedings of International Conference on Learning Representations (ICLR)}, 2021.

\end{thebibliography}
}


\end{document}


\maketitle

\section{Overview}



In this supplementary material, we provide additional details and more experimental results to further validate the proposed UniCD framework. The structure of this material is organized as follows.

\textbf{Details of the IRBFD Dataset.}
In Section 2, we describe the construction of the IRBFD dataset, including the synthetic process for generating nonuniformity bias fields in IRBFD-syn and the characteristics of real-world scenes in IRBFD-real. These datasets aim to comprehensively benchmark nonuniformity correction (NUC) and UAV detection methods.

\textbf{Further Validation Results of the UniCD Framework.}
In Section 3, we detail the training process of UniCD, including both separate and union training phases. We highlight the design of loss functions to ensure effective cooperation between the NUC and detection modules.

In Section 4, we present additional quantitative results that demonstrate the superior performance of UniCD compared to state-of-the-art methods. UniCD achieves high detection accuracy and robustness across synthetic and real-world datasets, even under challenging scenarios.

In Section 5, we provide more qualitative results to visually illustrate the advantages of UniCD. The visualizations emphasize its ability to correct severe nonuniformity while maintaining accurate target detection.

\textbf{Further Ablation Studies.}
In Section 6, we conduct ablation studies to evaluate the impact of key components, such as the degree of polynomial used in the NUC module and the inclusion of the TEBS and BR losses. These studies confirm the adaptability and effectiveness of UniCD under various configurations.

\textbf{Details of Metric Computation.}
In Section 7, we calculate the signal-to-clutter ratio gain (SCRG) to quantify the improvement in target detectability achieved by UniCD.

In Section 8, we compute the cosine similarity between feature maps to analyze the feature alignment introduced by different correction methods. This further supports the superiority of UniCD in enhancing feature representation for UAV detection.

This supplementary material highlights the robustness, efficiency, and effectiveness of the UniCD framework across a wide range of experimental scenarios, providing strong support for the claims made in the main paper.

\section{More Details about the Dataset IRBFD}
\subsection{Generation Process of the Synthetic Dataset IRBFD-syn }

\begin{figure*}[t]
\centering 
\includegraphics[width=6in,keepaspectratio]{supp_figs/different_k.pdf}
\caption{Visualization of degraded images with varying levels of nonuniformity degradation controlled by $k$. } 
\label{fig:impact_of_diffent_k} 
\end{figure*}

\textbf{Selection of Infrared Clear Images.}
To ensure the broad applicability and diversity of the dataset, we select the largest publicly available infrared UAV dataset \cite{2024TPAMIHuang} as the foundation. From this dataset, we uniformly and randomly sampled 30,000 images to serve as the infrared clear images. After sampling, we conduct a manual review to ensure that every image contains UAV targets. These images, along with their tracking annotations, are converted into the VOC dataset format for ease of use and compatibility with existing tools.

\begin{figure*}[t]
\centering 
\includegraphics[width=\textwidth]{supp_figs/background.pdf} %
\caption{Typical backgrounds in UAV surveillance scenarios from the IRBFD dataset.} 
\label{fig3:diversity_of_backgrounds} 
\end{figure*}

\textbf{Generation of Degraded Images with Nonuniformity Bias Fields.}
We model nonuniformity bias field using the following bivariate polynomial:
\begin{equation}
B(x_i, y_j) = \sum_{t=0}^{D} \sum_{s=0}^{D-t} a_{t,s}x_i^t y_j^s,
\label{sup:eq1}
\end{equation}
where coefficients $a_{t,s}$ are randomly setted to simulate varying bias levels. $D$ denotes the degree of the polynomial. By varying the degree $D$, we can obtain bias fields with different basis surfaces. In our work, we set the degree $D$ to 3. Then, by adding the bias field to a clear infrared image, we can obtain the degraded image with bias fields:
\begin{equation}
Y = C + k\ast B,
\end{equation}
where $ Y $, $ C $, and $B$ represent the degraded image, the clean image, and the bias field, respectively. $k$ is employed to control the severity of nonuniformity. The effects of different levels of degradation are shown in \cref{fig:impact_of_diffent_k}. When the degree of degradation is very low ($k$ is much smaller than 10), existing target detection methods can detection UAV targets without requiring correction. However, as the degradation becomes more severe ($k$ is much greater than 10), most existing correction methods fail and existing detection methods also struggle to detect the UAV targets. Therefore, we set $k$=10 in the dataset to evaluate the effectiveness of UniCD under challenging nonuniformity conditions.

\begin{figure}[t]
\centering 
\includegraphics[width=3.2in,keepaspectratio]{supp_figs/fig1.png}
\caption{Statistics of target scales in the IRBFD dataset.} 
\label{fig:scale_variation_of_targets} 
\end{figure}

\subsection{Statistics and Analysis of the IRBFD Dataset }
\textbf{Diversity of Backgrounds.} In \cref{fig3:diversity_of_backgrounds}, the first row presents typical background images from the synthetic dataset IRBFD-syn. For clarity of visualization, we display the infrared clear images at the top of \cref{fig3:diversity_of_backgrounds}. These scenes encompass two different lighting conditions (day and night), two seasons (autumn and winter), and a variety of backgrounds, including buildings (30\%), mountains (20\%), forests (5\%), urban areas (30\%), clouds (10\%), and water surfaces (3\%) \cite{2024TPAMIHuang}.

In the real dataset IRBFD-real, we collect UAV target data under various real-world bias field degradation scenarios. The dataset encompasses a diverse range of backgrounds, including dense clouds, trees, power lines, buildings, and farmlands, as shown at the bottom of \cref{fig3:diversity_of_backgrounds}. These real-world scenes were carefully selected to reflect practical environmental complexities, providing a comprehensive testbed for evaluating the robustness of the proposed method against real nonuniformity effects.

\textbf{Scale Variation of Infrared UAV Targets.} As shown in \cref{fig:scale_variation_of_targets}, in the IRBFD-syn dataset, tiny-scale UAV targets (Tiny, [2, 10)) account for approximately 37.8\% (9,852 targets), mini-scale UAV targets (Mini, [10, 20)) account for approximately 63.1\% (16,406 targets), small-scale UAV targets (Small, [20, 30)) account for approximately 11.9\% (3,110 targets), and medium \& normal-scale UAV targets (Medium \& Normal, [30, inf)) account for less than 2.5\% (632 targets).

In contrast, in the IRBFD-real dataset, tiny-scale UAV targets account for approximately 25.8\% (5,383 targets), mini-scale UAV targets account for approximately 68.8\% (14,345 targets), small-scale UAV targets account for only 1.7\% (356 targets), and medium \& normal-scale UAV targets account for less than 0.2\% (36 targets).

The real-world dataset, IRBFD-real, closely reflects the challenges of real-world anti-UAV scenarios. The significant proportion of tiny and mini UAV targets highlights the difficulty of detecting smaller targets, making it more representative of real-world application needs.


\begin{figure}[t]
\centering 
\includegraphics[width=2.5in,keepaspectratio]{supp_figs/Position_Distribution.pdf}
\caption{Position distribution of targets in the IRBFD dataset.} 
\label{fig2:position_distribution_of_targets} 
\end{figure}

\textbf{Comprehensiveness of UAV Target Position Distribution.} \Cref{fig2:position_distribution_of_targets} illustrates the position distribution of UAV targets within the IRBFD dataset, including both the synthetic dataset (IRBFD-syn) and the real-world dataset (IRBFD-real).

The UAV targets in the IRBFD dataset are comprehensively distributed across the entire image space, as shown in both subfigures. This design ensures that the datasets cover a wide range of spatial configurations, providing diverse scenarios for evaluating nonuniformity correction (NUC) and UAV detection methods. The distribution of targets reflects a balanced dataset that supports robust model training and testing across varying environmental conditions.




\section{More Training Details about the UniCD}
When training UniCD on the IRBFD-syn dataset, we initially train the NUC module and the infrared UAV target detection module separately. During the training of the NUC module, the loss function is defined as shown in \cref{main-sec:method}, \cref{main-eq5} of the main text. For the infrared UAV target detection module, the loss function is defined as shown in \cref{main-eq8} of the main text. Once both sub-modules are sufficiently trained, we proceed to train the union framework. In this phase, the loss function is given by \cref{main-eq11} of the main text, where the BR loss is designed to balance the conflict between the NUC module and the UAV target detection module.

On the IRBFD-real dataset, we pre-train only the UAV target detection module, while the NUC module retains the weights trained on the IRBFD-syn dataset without further updates. During union training, the NUC module is frozen, and the loss function defined in \cref{main-eq11} of the main text is used for optimization.

\section{More Quantitative Comparisons of UniCD with Existing Methods}
\begin{table}[t]
\centering
\setlength\tabcolsep{0.8pt} 
\fontsize{7}{9.2}\selectfont  
\setlength{\abovecaptionskip}{3pt} 
\caption{Quantitative comparison of the proposed method with SOTA methods on the synthetic dataset IRBFD-syn. For the separation strategy, each correction method corresponds to multiple different detection methods. {\textbf{Bold}} and \underline{underline} indicate the best and the second best results, respectively.}
\begin{tabular}{c|ccc|ccccc} 
\hline 
\multirow{2}{*}{Strategy} & \multicolumn{3}{c|}{Module\vspace{-.3em} }  & \multicolumn{5}{c}{Metrics} \\ \cline{2-4}  \cline{5-9}
& NUC & Detection & Pub'Year  & PSNR $\uparrow$ & SSIM $\uparrow$ & P $\uparrow$ & R $\uparrow$ & FPS $\uparrow$ \\  \hline
\multirow{6}{*}{Direct} & \multirow{6}{*}{-} & Deformable DETR & ICLR'21  & \multirow{6}{*}{-} & \multirow{6}{*}{-} & 0.614 & 0.630 & 24 \\
 &  & DINO & ICLR'23   &  &  & 0.904 & {0.640} & 26 \\
 &  & DAGNet & TII'23   &  &  & \underline{0.994} & 0.635 & \textbf{43} \\
 &  & LESPS & CVPR'23   &  &  & 0.033 & 0.446 & 12 \\
 &  & MSHNet & CVPR'24   &  &  & 0.407 & 0.421 & 41 \\ 
 &  & YOLO11L & 2024   &  &  & 0.963 &0.602 & \underline{42} \\   \hline
 \multirow{20}{*}{Separate} & \multirow{4}{*}{Liu} &DINO & \multirow{4}{*}{IPT'16} & \multirow{4}{*}{16.800} & \multirow{4}{*}{0.8289}  &0.868 & 0.599 & \textless1 \\
& &MSHNet &  &  &  & 0.686 & \underline{0.663} & \textless1 \\
& & YOLO11L  & & & & 0.898 & 0.574 & \textless1 \\
 &  & DAGNet   &  &  &  & 0.978 & 0.578 & \textless1 \\ \cline{2-9}
 & \multirow{4}{*}{DMRN} 
 & DINO & \multirow{4}{*}{GRSL'19}  & \multirow{4}{*}{\underline{24.467}} & \multirow{4}{*}{\underline{0.8600}}  &0.841 & 0.585 & 23 \\
& & MSHNet &  &  &  & 0.669 & 0.663 & 35 \\ 
& & YOLO11L & & & & 0.923 & 0.550 & 35 \\
 &  & DAGNet   &  &  &  & 0.966 & 0.595 & 36 \\  \cline{2-9}
 & \multirow{4}{*}{Shi}  
 & DINO & \multirow{4}{*}{AO'22}  & \multirow{4}{*}{13.974} & \multirow{4}{*}{0.7783}  &0.813 & 0.557 & \textless1 \\
& & MSHNet &  &  &  & 0.584 & 0.618 & \textless1 \\
& & YOLO11L  & & & & 0.924 & 0.455 & \textless1 \\
 &  & DAGNet   &  &  &  & 0.966 & 0.472 & \textless1 \\  \cline{2-9}
 & \multirow{4}{*}{TV-DIP}
 & DINO & \multirow{4}{*}{IPT'23}   & \multirow{4}{*}{13.397} & \multirow{4}{*}{0.6374}  &0.086 & 0.115 & 21 \\
 & & MSHNet &  &  &  & 0.078 & 0.077 & 29 \\
 & & YOLO11L  & & & & 0.131 & 0.020 & 29 \\
 &  & DAGNet   &  &  &  & 0.599 & 0.020 & 30 \\  \cline{2-9}
 & \multirow{4}{*}{AHBC} 
  & DINO & \multirow{4}{*}{TGRS'24}   & \multirow{4}{*}{13.954} & \multirow{4}{*}{0.6763} &0.417 & 0.294 & \textless1 \\
  & & MSHNet &  &  &  & 0.300 & 0.208 & \textless1 \\
 & & YOLO11L & & & & 0.825 & 0.080 & \textless1 \\
 &  & DAGNet   &  &  &  & 0.724 & 0.040 & \textless1 \\  \hline
Union & \multicolumn{2}{c}{UniCD} & -  & \textbf{31.961} & \textbf{0.9827} & \textbf{0.999} & \textbf{0.822} & 32 \\ \hline
\end{tabular} 
\label{tab:irbfd_simulation_sup}
\end{table}
\begin{table}[t]
\centering
\setlength\tabcolsep{2pt} 
\fontsize{8}{9}\selectfont  
\setlength{\abovecaptionskip}{3pt} 
\caption{Quantitative comparison of the proposed method with SOTA methods on the real dataset IRBFD-real. {\textbf{Bold}} and \underline{underline} indicate the best and the second best results, respectively. }
\begin{tabular}{ccccccc} \hline
Strategy & NUC & Detection & SCRG $\uparrow$ & P $\uparrow$ & R $\uparrow$ \\   \hline
\multirow{3}{*}{Direct} & \multirow{3}{*}{-} & DINO & \multirow{3}{*}{-} & 0.971 & 0.660 \\
 &  & YOLO11L &  & 0.966 & 0.843 \\
 &  & DAGNet &  & \underline{0.992} & \underline{0.871} \\   \hline
\multirow{9}{*}{Separate} &  
\multirow{3}{*}{TV-DIP}   & DINO & \multirow{3}{*}{0.412} & 0.094 & 0.083 \\
& & YOLO11L &  & 0.521 & 0.024 \\
& & DAGNet &  & 0.663 & 0.026 \\  \cline{2-6}
& \multirow{3}{*}{DMRN}   & DINO & \multirow{3}{*}{0.997} & 0.687 & 0.480 \\
 & & YOLO11L &  & 0.918 & 0.296 \\
 &  & DAGNet &  & {0.929}& 0.345 \\  \cline{2-6}
  & \multirow{3}{*}{AHBC}   & DINO & \multirow{3}{*}{\underline{1.146}} & 0.964 & 0.649 \\
 & & YOLO11L &  & 0.940 & 0.633 \\
 &  & DAGNet &  & {0.986}& 0.699 \\   \hline
 Union & \multicolumn{2}{c}{UniCD} & \textbf{1.286} & \textbf{0.994} & \textbf{0.901} \\  \hline
\end{tabular}
\label{tab:irbfd_real_sup} 
\end{table}

We further expand the experiments on the synthetic dataset IRBFD-syn and the real-world dataset IRBFD-real, with the results presented in \cref{tab:irbfd_simulation_sup} and \cref{tab:irbfd_real_sup}, to verify the superior performance of the proposed UniCD framework.

On the IRBFD-syn dataset, we evaluate additional target detection methods (DINO \cite{2023ICLRZhang}  and MSHNet \cite{2024CVPRLiu}) under separate strategies. The experiments demonstrate that, among the separate strategies, DL-driven correction methods (e.g., DMRN \cite{2019GRSLChang}  ) significantly improve image quality and achieve notable PSNR and SSIM results. However, their detection performance is limited due to the independent handling of correction and detection modules, resulting in low recall. Traditional model-driven correction methods (e.g., Liu \cite{2016IPTLiu} , Shi \cite{2022AOShi} , and AHBC \cite{2024TGRSXie} ) struggle to handle severe degradation scenarios, further impairing detection accuracy. In contrast, UniCD achieves a PSNR of 31.961 and an SSIM of 0.9827 while maintaining outstanding detection precision (P = 0.999) and recall (R = 0.822), showcasing its ability to handle image correction and target detection simultaneously in a unified framework.

On the IRBFD-real dataset, we expand the analysis with more NUC methods and target detection combinations under a separate processing strategy. The experiments reveal that separate strategies, such as TV-DIP \cite{2023IPTLiu}  and AHBC, face significant limitations on real-world data. For instance, TV-DIP often deteriorates the image content, resulting in extremely low detection precision and recall. While AHBC performes better in certain scenarios, its overall precision and recall still fell short compared to UniCD. In contrast, UniCD consistently achieves a SCRG of 1.286 and surpassed all other methods with a precision of P = 0.994 and a recall of R = 0.901. These results highlight UniCD's adaptability and robustness in handling diverse real-world nonuniformity scenarios.

Furthermore, UniCD's real-time processing capability (32 FPS) ensures its practical deployment even under resource-constrained conditions. The experimental results further confirm that UniCD provides a comprehensive solution to address the challenges posed by nonuniformity in infrared UAV detection tasks.

\section{More Qualitative Comparisons of UniCD with Existing Methods}
We evaluate more correction-then-detection methods on both synthetic and real-world datasets across diverse scenarios, including buildings, hillsides, clouds, forests, and urban regions. These visualizations further validate the findings from the quantitative experiments presented in the main paper.

\Cref{fig-syn} presents the qualitative results on the synthetic dataset. The analysis is supplemented by including DINO and MSHNet as detection methods applied to various NUC correction approaches. The results indicate that traditional NUC methods, such as Liu, TV-DIP, and AHBC, paired with DINO or MSHNet, often fail to effectively handle severe nonuniformity effects:
\begin{itemize}
    \item Liu + DINO/MSHNet results in extremely low detection confidence in most scenarios (e.g., 0.02 confidence or undetected targets in hillside scenes) due to insufficient correction of the bias field.
    \item TV-DIP + DINO/MSHNe\textbf{t} fails to restore sufficient target features, particularly in forested or cluttered environments, leading to missed detections.
    \item AHBC + DINO/MSHNet, while slightly more effective, still produces false positives or weak  confidence scores.
\end{itemize}

\begin{figure*}[pht!]
\centering 
\includegraphics[width=6in,keepaspectratio]{supp_figs/syn-detect-zip.pdf} 
\caption{Visual comparison of results from separate correction-then-detection methods and our UniCD on the synthetic dataset IRBFD-syn .
Closed-up views are shown in the left bottom corner. Boxes in green and red represent ground-truth and correctly detected targets, respectively.} 
\label{fig-syn} 
\end{figure*}

In contrast, our UniCD consistently outperforms all these combinations, achieving clear corrected images and accurate UAV target detection with high confidence scores (1.00 across all scenarios). This superior performance is attributed to the robust integration of parametric modeling in the NUC module and the auxiliary loss in the detection backbone, enabling accurate parameter estimation and enhanced feature representation under severe nonuniformity conditions.
\begin{figure*}[ht!]
\centering 
\includegraphics[width=\textwidth]{supp_figs/real.pdf}
\caption{Visual comparison of results from separate correction-then-detection methods and our UniCD on the real dataset IRBFD-real.} 
\label{fig-real} 
\end{figure*}

\Cref{fig-real} shows the qualitative results on the real-world dataset. We further add the following content: (1) DINO as the detection network applied to different NUC correction methods. (2) DMRN as a newly introduced NUC method, combined with three detection methods: DINO, YOLO11L, and DAGNet.
The results demonstrate the limitations of conventional NUC methods when applied to real-world data:
\begin{itemize}
    \item TV-DIP + DINO fails to effectively correct the bias field and instead degrades image content, resulting in extremely low detection confidence scores or entirely missed targets.
    \item DMRN + DINO, while improving bias field correction, introduces block artifacts that impair detection accuracy. DMRN + YOLO11L \cite{2024YOLO11Jocher} and DMRN + DAGNet \cite{2023TIIFang} show moderate improvements but struggle to maintain consistent detection in complex urban or cloudy scenes.
    \item AHBC + DINO performs inconsistently, with detection confidence dropping as low as 0.36 in dense cloud scenarios.
\end{itemize}

In contrast, our UniCD provides consistently superior results across all scenarios, achieving high detection confidence scores (1.00) even in challenging conditions, such as dense clouds and cluttered urban environments. These results highlight UniCD's ability to simultaneously balance correction and detection tasks while effectively addressing complex real-world nonuniformity effects.

Overall, these supplementary visualization results demonstrate the superiority of UniCD compared to conventional approaches in both synthetic and real-world settings. This further reinforces UniCD's capability to address challenging nonuniformity effects while delivering robust and reliable UAV detection performance.

\begin{figure*}[ht!]
\centering 
\includegraphics[width=6.8in,keepaspectratio]{supp_figs/base_surface-3.pdf} 
\caption{Visualization of 2D and 3D basis surfaces for polynomials of different degrees. The surface sets for the second-degree, third-degree, and fourth-degree models are highlighted with yellow, green, and blue background boxes, respectively. We employ blue and red boxes to represent similar redundant surfaces for a degree equal to 4. } 
\label{fig:2D3D_basis_surface}   
\end{figure*}

\section{Further Ablation Studies}
\subsection{Impact of Polynomial Degree on the NUC Module}
In this section, we provide a detailed discussion on the rationale for selecting the third-degree polynomial model.

\textbf{Definition of the Polynomial  Basis Surfaces. } As presented in \cref{sup:eq1}, the polynomial model \( B(x_i, y_j) \) serves as an effective representation of a surface in a two-dimensional space, where each term \( x_i^t  y_j^s \) represents a basis surface that contributes to the overall shape of the surface. The influence of each basis surface is modulated by its corresponding coefficient \( a_{t,s} \), which determines the weight and impact of that component. By appropriately adjusting these coefficients, the polynomial model can capture complex spatial variations and transformations, making it well-suited for modeling nonuniformity bias fields. This structure representation strikes a balance between expressiveness and computational efficiency, providing a flexible yet compact framework for accurately representing diverse patterns of bias field distortions.

\textbf{Why Choose the Third-degree Polynomial? }As shown in \cref{fig:2D3D_basis_surface}, the comparison of 2D basis surfaces and their corresponding 3D visualizations across different polynomial degrees highlights key considerations for selecting the optimal degree. Higher-degree polynomials inherently encompass the basis surfaces of lower-degree ones, which indicates that very low-degree polynomials lack sufficient expressive power to model complex spatial variations in the bias field effectively. However, as the degree increases beyond three, redundant basis surfaces emerge (highlighted in red boxes in the figure), introducing overlapping representations. This redundancy not only diminishes modeling efficiency but also increases the risk of overfitting and numerical instability.

In addition, higher-degree polynomials require estimating a larger number of coefficients, which complicates the optimization process and slows convergence during training. In contrast, the third-degree polynomial strikes an ideal balance by offering adequate expressive power while maintaining computational simplicity and stability, making it the optimal choice for our framework.

Furthermore,  the  quantitative results in \cref{main-tab:degree_selection} of the main text also demonstrate that the third-degree polynomial achieves the best performance among all configurations. A second-degree polynomial, with only 6 coefficients, struggles to model the bias field accurately, resulting in poor correction performance (PSNR = 13.5279, SSIM = 0.7590) and low recall (R = 0.433). Increasing the degree to three improves both correction quality (PSNR = 39.050, SSIM = 0.9970) and detection performance (R = 0.810), achieving the best balance between accuracy and complexity. Polynomials of higher degrees (e.g., 4 and 5) introduce more coefficients, leading to a slight decline in correction performance and recall (e.g., PSNR = 31.744, R = 0.788 for degree 4). This indicates diminishing returns and potential risks of overfitting.




\subsection{Impact of Different Levels of Degradation on UniCD}
\begin{table}[t]
\centering
\fontsize{9}{10}\selectfont  
\setlength{\abovecaptionskip}{3pt} 
\caption{Robustness analysis of UniCD across different levels of nonuniformity degradation.}
\setlength{\belowcaptionskip}{-0.2cm}
\begin{tabular}{ccccccc} \hline
$k$ & PSNR & SSIM & P&R \\   \hline
3 & 29.119 & 0.9891 & 0.998&0.820 \\  
5 & 34.907 & 0.9950 & 0.998&0.820 \\  
12 & 38.361 & 0.9968 & 0.998&0.820  \\  \hline
\end{tabular}
\label{tab:kvalue-sup}
\end{table}

To analyze the adaptability of our proposed UniCD framework, we conduct experiments on the IRBFD-syn dataset by training the model on images with a fixed degradation level (\( k=10 \)) and testing it across varying levels of degradation (\( k=3, 5, 12 \)). The results, summarized in \cref{tab:kvalue-sup} demonstrate the robustness of our method under different nonuniformity conditions.  

Specifically, the PSNR and SSIM values increase consistently with higher \( k \)-values, indicating that the bias field correction module effectively reconstructs clearer images as the degradation severity rises. This improvement reflects the model's ability to handle more challenging conditions with greater nonuniformity intensity.  

Despite the variation in image quality, the detection performance metrics (P and R) remain consistent across all tested \( k \)-values. Both precision (P = 0.998) and recall (R = 0.820) are maintained at high levels, showcasing UniCD's ability to balance correction and detection tasks effectively. This result highlights the flexibility and generalization capability of the framework, ensuring reliable performance even when deployed under varying real-world degradation scenarios.  

These findings confirm that UniCD is not only robust to different nonuniformity levels but also achieves stable detection results, reinforcing its practical applicability in diverse environmental conditions.

\subsection{Number of Parameters and Computational Complexity of the NUC Module}
\begin{table}[t]
\centering
\fontsize{9}{10}\selectfont  
\caption{Comparison of number of parameters and computational complexity for different NUC methods.}
\setlength{\belowcaptionskip}{-0.2cm}
\begin{tabular}{ccc} \hline
Methods & Params (M) & FLOPs (G) \\   \hline 
DMRN & 0.7398 & 76.2524  \\  
TV-DIP& 2.1817 & 75.3930 \\   
Our NUC & 0.3966 & 1.6809   \\   \hline 
\end{tabular}
\label{tab:param}
\end{table}

\Cref{tab:param} gives a comparison of number of parameters and computational complexity (measured in FLOPs) among different NUC methods, including DMRN, TV-DIP, and our proposed NUC module. They are all methods based on deep learning. The results demonstrate the significant advantages of our NUC method in terms of both model efficiency and computational complexity.

\begin{itemize}
    \item \textbf{Number of Parameters.}  
    Our NUC module requires only 0.3966 million parameters, which is approximately 46.4\% smaller than\textbf{ }DMRN (0.7398M) and 81.8\% smaller than TV-DIP (2.1817M). This reduction in number of parameters directly translates to lower memory usage and faster runtime performance, making it more suitable for resource-constrained applications.
    
    \item \textbf{Computational Complexity.}  
    The proposed NUC module achieves a significant reduction in Floating Point Operations (FLOPs), requiring only 1.6809G, which is 55 times smaller than DMRN (76.2524G) and 44 times smaller than TV-DIP (75.3930G). Such a significant reduction indicates that our module is highly optimized for real-time processing, further enhancing its practical deployment capabilities.
\end{itemize}

In summary, the results demonstrate that our NUC module not only achieves superior nonuniformity correction performance but also does so with minimal computational overhead, significantly outperforming other methods in terms of efficiency. These attributes make our method an ideal solution for real-world scenarios where computational resources are limited.

\subsection{Impact of the TEBS Loss on the Detection Module}
In this section, we further demonstrate the superiority of the TEBS loss, highlighting its effectiveness in improving training performance, as well as enhancing target features while suppressing background. 
\begin{figure}[t]
\centering 
\includegraphics[width=3.2in,keepaspectratio]{supp_figs/TEBS_loss.png}
\caption{Comparation of the localization and classification loss curves during training without and with the TEBS loss.} 
\label{fig:tebsloss} 
\end{figure}

\Cref{fig:tebsloss} illustrates the comparison of the localization and classification loss curves during training without and with the TEBS loss. The upper plot shows the location loss, while the lower plot represents the classification loss. It is evident that with the TEBS loss (solid orange line), both the location and classification losses converge more smoothly and to significantly lower values compared to training without the TEBS loss (dashed blue line). Specifically, the TEBS loss effectively stabilizes the training process, reduces oscillations, and enhances convergence speed, demonstrating its clear advantage in improving both localization and classification performance.

\begin{figure*}[t]
\centering 
\includegraphics[width=6.1in,keepaspectratio]{supp_figs/TEBS.pdf}
\caption{Comparison of feature maps from different stages of the detection backbone without and with the TEBS loss. The green dashed circles and 
solid circles represent the target enhancement effects without and with the TEBS loss, respectively. The red dashed circles represent the residual background.} 
\label{fig:tebsfeat} 
\end{figure*}

\Cref{fig:tebsfeat} shows the comparison of feature maps from different stages of the detection backbone without and with the TEBS loss. The TEBS loss imposes explicit supervision on the target and background masks during training, which significantly enhances the feature representation of UAV targets while suppressing irrelevant background information.

From Stage 1 to Stage 4, it can be observed that without the TEBS loss, the feature maps exhibit noticeable residual background interference (highlighted by red circles), especially in complex scenes such as urban areas or dense vegetation. This interference not only reduces the distinguishability of the UAV target but also increases the likelihood of false positives or missed detections. In contrast, with the TEBS loss, the feature maps progressively refine the representation of the UAV targets, as indicated by the high-intensity regions (green circles) becoming more focused and distinct. The background clutter is effectively suppressed, particularly in later stages of the backbone network.

This improvement highlights the contribution of the TEBS loss in enforcing spatial and semantic constraints on the backbone features, enabling the network to prioritize target regions while mitigating distractions from irrelevant features. As a result, the detection performance of the framework is substantially enhanced, as corroborated by the quantitative results and visualizations provided in the main text and supplementary material.

\subsection{Impact of the BR Loss on the Union Framework}
\begin{table}[t]
\centering
\setlength\tabcolsep{1.2pt} 
\fontsize{8.5}{10}\selectfont  
\setlength{\abovecaptionskip}{1pt} 
\caption{Impact of different union training methods on performance.}
\begin{tabular}{ccccccccc} \hline
\multirow{2}{*}{ Row}&\multirow{2}{*}{ \makecell{Correction\\loss}}&\multirow{2}{*}{ \makecell{Frozen\\NUC}} &\multirow{2}{*}{\makecell{Detection\\loss}}& \multirow{2}{*}{\makecell{BR\\loss}}& \multicolumn{4}{c}{Metrics} \\  \cline{6-9}
 &&&&& PSNR & SSIM & P & R \\ \hline
 1& $\surd$&$\times$& $\surd$ & $\times$  & 33.024 & 0.9827  &0.998   & 0.791 \\
2 & $\times$& $\times$&$\surd$ & $\times$ & 17.940 & 0.8910 &  0.989& 0.811\\
3 & $\times$&$\surd$& $\surd$ & $\times$ & \textbf{37.722 } & \textbf{0.9960} &0.998  & 0.804\\
4& $\times$&$\times$& $\surd$ & $\surd$ & 31.961 & 0.9827 & \textbf{0.999} & \textbf{0.822} \\  \hline
\end{tabular}
\label{tab:union_ablation_syn}
\end{table}
\begin{figure*}[h]
\centering 
\includegraphics[width=5in,keepaspectratio]{supp_figs/detectionfriend.pdf}
\caption{Visualization comparison of feature maps across four stages of the detection backbone under separate and union training. The green and red dashed circles represent the enhanced target region and the significant residual background, respectively.} 
\label{fig:detectinoFriend} 
\end{figure*}
In this section, we focus on discussing the impact of various union training methods on the relationship between the NUC and target detection sub-tasks.
\Cref{tab:union_ablation_syn} presents a comparison of four different union training strategies for combining the correction and detection modules, highlighting the impact of the BR loss in the UniCD framework.

(1) \textit{Training with the correction and detection losses} (Row 1).
In this approach, the correction and detection losses are directly added to jointly train the two modules. However, since the optimization objectives of the two losses differ, this method introduces conflicts that prevent either module from achieving optimal performance. As a result, both correction and detection metrics suffer from suboptimal outcomes.

(2) \textit{Training with the detection loss only }(Row 2).
This strategy uses only the detection loss to train both the correction and detection modules. Due to the lack of supervision on the correction module, its performance degrades significantly, leading to a sharp decline in PSNR and SSIM. Although the detection metrics improve slightly (e.g., Recall = 0.811), the severe degradation in correction quality makes this method unsuitable for scenarios requiring high-quality visual outputs.

(3) \textit{Training with the detection loss and the frozen correction module} (Row 3).
In this method, the pre-trained correction module is frozen, and only the detection module is updated during training. While this approach achieves the best correction metrics (PSNR = 37.722, SSIM = 0.9960), the lack of interaction between the modules means that the correction module does not retain information beneficial to detection. Consequently, the detection metrics decline, with Recall dropping to 0.804.

(4) \textit{Training UniCD with the detection loss and the BR loss} (Row 4).
Our UniCD framework introduces the BR loss to enforce feature-level supervision from the detection backbone on the correction module. This self-supervised mechanism ensures that the correction module not only maintains visual quality (PSNR = 31.961, SSIM = 0.9827) but also aligns with the detection objectives. By jointly training both modules, the detection module can guide the correction module to produce feature maps that enhance UAV detection performance, achieving the highest detection metrics (P = 0.999, R = 0.822). This demonstrates the effectiveness of the BR loss in balancing the objectives of correction and detection while achieving superior overall performance.

\subsection{Detection-Friendliness of the NUC Module under Union Training}
 In this section, we verify that the proposed union training method enables the NUC module to produce detection-friendly results, which are reflected in the feature maps of the detection backbone. As shown in \cref{fig:detectinoFriend}, we present the feature maps across four stages of the detection backbone under separate and union training strategies: the separate training approach (first row and third row) and the union training approach (second row and fourth row).

In the separate training approach, the NUC module and the detection network are trained independently. As shown in the feature maps from the separate training approach, the extracted features exhibit scattered and noisy activations, with limited focus on the UAV target regions. This indicates a lack of alignment between the correction and detection tasks, leading to suboptimal feature representations for the detection network.

In contrast, the union training approach integrates the NUC module and detection network into a unified end-to-end framework. As observed in the second row and fourth row, the feature maps from union training display more concentrated and structured activations around the UAV target regions. This improvement highlights the detection-friendliness of the features produced by the NUC module under joint optimization, effectively suppressing irrelevant background information and enhancing the features critical for accurate UAV detection.

This comparison demonstrates that union training not only improves the compatibility between the NUC and detection modules but also enables the NUC module to produce features that are more beneficial for downstream detection tasks, resulting in improved overall performance.

\section{Calculation of Signal-to-Clutter Ratio Gain (SCRG)} \label{sec:SCRG}
\begin{figure}[t]
\centering 
\includegraphics[width=1.6in,keepaspectratio]{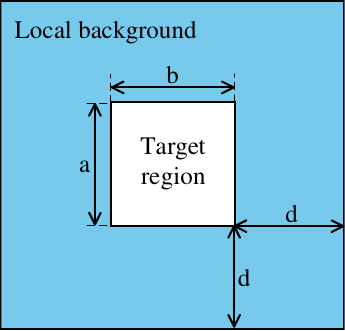}
\caption{The bounding box of a small target and the adjacent background box.} 
\label{figSCR} 
\end{figure}

The signal-to-clutter ratio (SCR) is utilized to measure the difficulty of target detection in a local region, can be calculated by:
\begin{equation}
\text{SCR} = \frac{|\mu_t - \mu_b|}{\sigma_b},
\end{equation}
where $\mu_t$ and $\mu_b$  represent the average pixel values of the target region and the surrounding neighboring region, respectively. $\sigma_b$ is the standard deviation of the pixel values in the surrounding neighboring region of the target. As shown in \cref{figSCR}, we assume the size of the small UAV target is $a\times b$, and then the size of its background region is $(a + 2d) \times (b + 2d)$, where $d$ is the pixel width of neighboring area. We set $d = 5$ pixels in our experiment.

The signal-to-clutter ratio gain (SCRG) is the ratio of SCR in the corrected image to that in the original image, used to evaluate the improvement in target detectability achieved by the correction method, which can be defined as:
\begin{equation}
\text{SCRG} = \frac{\text{SCR}_\text{out}}{\text{SCR}_\text{in}}.
\end{equation}

\section{Calculation of the Cosine Similarity Between Two Feature Maps}  \label{sec:cos_sim}
To compute the cosine similarity between two feature maps $ A $ and $ B $, we assume they have the same shape, $ m \times n $ (i.e., $ A $ and $ B $ each have $ m $ rows and $ n $ columns). The cosine similarity between feature maps can be interpreted as the cosine similarity between each pair of row vectors from $ A $ and $ B $, resulting in an $ m \times m $ similarity matrix.

\subsection{Definition of Cosine Similarity}
The cosine similarity between two vectors $ \mathbf{a} $ and $ \mathbf{b} $ is defined as:
\begin{equation}
\text{Cos\_Sim}(\mathbf{a}, \mathbf{b}) = \frac{\mathbf{a} \cdot \mathbf{b}}{\|\mathbf{a}\|_2 \|\mathbf{b}\|_2},
\end{equation}
where:
\begin{itemize}
    \item $ \mathbf{a} \cdot \mathbf{b} = \sum_{k=1}^n a_k b_k $ is the dot product of $ \mathbf{a} $ and $ \mathbf{b} $.
    \item $ \|\mathbf{a}\|_2 = \sqrt{\sum_{k=1}^n a_k^2} $ and $ \|\mathbf{b}\|_2 = \sqrt{\sum_{k=1}^n b_k^2} $ are the $ L_2 $ norms of $ \mathbf{a} $ and $ \mathbf{b} $, respectively.
\end{itemize}

\subsection{Computing the Dot Product Matrix}
For two feature maps $ A $ and $ B $, let $ A_i $ and $ B_j $ represent the $ i $-th row of $ A $ and the $ j $-th row of $ B $, respectively. We can construct a dot product matrix $ D $, where each element $ D_{ij} $ represents the dot product of row $ A_i $ with row $ B_j $:
\begin{equation}
D_{ij} = A_i \cdot B_j = \sum_{k=1}^n A_{ik} B_{jk}.
\end{equation}

Thus, the matrix $ D $ can be expressed as:
\begin{equation}
D = A B^T.
\end{equation}

\subsection{Calculating the $ L_2 $ Norms of Row Vectors}
For each row $ A_i $ of feature map $ A $ and each row $ B_j $ of feature map $ B $, we compute their $ L_2 $ norms:
\begin{equation}
\|A_i\|_2 = \sqrt{\sum_{k=1}^n A_{ik}^2} \quad \text{and} \quad \|B_j\|_2 = \sqrt{\sum_{k=1}^n B_{jk}^2}.
\end{equation}

The norms for all rows of $ A $ and $ B $ can be represented as column vectors:
\begin{equation}
\|A\|_2 = \begin{bmatrix} \|A_1\|_2 \\ \|A_2\|_2 \\ \vdots \\ \|A_m\|_2 \end{bmatrix}, \quad \|B\|_2 = \begin{bmatrix} \|B_1\|_2 \\ \|B_2\|_2 \\ \vdots \\ \|B_m\|_2 \end{bmatrix}.
\end{equation}

\subsection{Forming the Outer Product of Norms}
Using the $ L_2 $ norms $ \|A\|_2 $ and $ \|B\|_2 $, we construct an $ m \times m $ matrix $ N $, where each element $ N_{ij} $ represents the product of the norms:
\begin{equation}
N_{ij} = \|A_i\|_2 \|B_j\|_2.
\end{equation}

Therefore, the matrix $ N $ can be represented as:
\begin{equation}
N = \|A\|_2 \|B\|_2^T.
\end{equation}

\subsection{Calculating the Cosine Similarity Matrix}
Finally, we compute the cosine similarity matrix by dividing each element of the dot product matrix $ D $ by the corresponding element in the norm product matrix $ N $:
\begin{equation}
\text{Cos\_Sim}_{ij} = \frac{D_{ij}}{N_{ij}} = \frac{A_i \cdot B_j}{\|A_i\|_2 \|B_j\|_2}.
\end{equation}

In matrix form, the cosine similarity between feature maps $ A $ and $ B $ is given by:
\begin{equation}
\text{Cos\_Sim}(A, B) = \frac{A B^T}{\|A\|_2 \|B\|_2^T}.
\end{equation}

\subsection{Final Formula Summary}
Thus, the cosine similarity matrix can be expressed as:
\begin{equation}
\text{Cos\_Sim}(A, B) = \frac{A B^T}{\sqrt{\sum_{k=1}^n A_{ik}^2} \cdot \sqrt{\sum_{k=1}^n B_{jk}^2}}.
\end{equation}

This provides the cosine similarity for each pair of row vectors in $ A $ and $ B $.

\bibliographystyle{plain} 
\bibliography{sup_ref} 